\documentclass[preprint]{imsart}

\RequirePackage[OT1]{fontenc}
\RequirePackage{amsthm,amsmath}
\RequirePackage[numbers]{natbib}
\RequirePackage[colorlinks,citecolor=blue,urlcolor=blue]{hyperref}
\RequirePackage{hypernat}

\usepackage{color}
\usepackage{fancybox}
\usepackage{algorithmic}
\usepackage{algorithm}
\usepackage{graphicx}
\usepackage{subfigure}
\usepackage{verbatim}
\usepackage{url}
\usepackage{bbm}
\usepackage{amssymb}
\usepackage{amsfonts}

\newcommand{\indicator}[1]{\mathbbm{1}_{\left[ {#1} \right] }}

\startlocaldefs
\numberwithin{equation}{section}
\theoremstyle{plain}

\endlocaldefs

\begin{document}

\begin{frontmatter}
\title{Mixed-Membership Stochastic Block-Models for Transactional Networks\protect}
\runtitle{Transactional Mixed-Membership Stochastic Block-Model}

\begin{aug}
\author{\fnms{Mahdi} \snm{Shafiei}\ead[label=e1]{mahdi.shafiei@acadiau.ca}}
\and
\author{\fnms{Hugh} \snm{Chipman}\ead[label=e2]{hugh.chipman@acadiau.ca} \ead[label=u1,url]{http://math.acadiau.ca/chipmanh/}}

\runauthor{M. Shafiei, H. Chipman}

\affiliation{Acadia University}

\address{Department of Mathematics \& Statistics, Acadia University\\
12 University Avenue, Huggins Science Hall, Wolfville, NS  Canada  B4P 2R6\\ 
\printead{e1}
\phantom{E-mail:\ }\printead*{e2}}
\end{aug}

\begin{abstract}
Transactional network data can be thought of as a list of one-to-many communications
(e.g., email) between nodes in a social network.
Most social network
models convert this type of data into binary relations between pairs
of nodes.  We develop a latent mixed membership model capable of modeling
richer forms of transactional network data, including relations between
more than two nodes. The model can cluster nodes and predict transactions. 
The block-model nature of the model implies that groups can be
characterized in very general ways.  
This flexible notion of group structure enables discovery of rich structure
in transactional networks.
Estimation and inference are accomplished via a variational
EM algorithm. Simulations indicate that the learning algorithm can recover
the correct generative model.  Interesting structure is discovered in
the Enron email dataset and
another dataset extracted from the Reddit website. Analysis of the Reddit data is facilitated by a novel performance measure for comparing two soft clusterings. 
The new model is superior at discovering mixed membership in groups and in
predicting transactions.
\end{abstract}

\begin{keyword}
\kwd{Social Network Analysis}
\kwd{Clustering; Mixed-membership}
\kwd{Variational EM} 
\kwd{Email Data}
\end{keyword}

\end{frontmatter}

\section{Introduction}

With the popularity of online social networks, discussion forums and
widespread use of electronic means of communication including email
and text messaging, the study of network-structured data has become quite
popular.  

Social network data typically consist of a group of {\em nodes} (or actors) and a
list of {\em relations} between nodes.  The most common models assume that
relations occur between pairs of nodes, and that a relation takes a binary
value (presence/absence).  Such data can be conceptualized as a graph, and
analogously, relations can be directed or undirected.  A canonical example of
such data would be a group of people (nodes) and friendship relations
between them.  If each person identifies their friends, then the friendship
relation can be directional ($A$ likes $B$ but $B$ does not like $A$).  

The assumptions that relations are binary-valued and occur between pairs of nodes do not always hold 
for network data.  In many cases, the data are {\em transactional},
with multiple instances of communication between individuals occurring
over time. For example, with telephone calls, a pair of nodes is
involved in a call but the relation is {\em transactional} (i.e. a list of
calls), rather than being binary-valued.  In email data, relations are
transactional and can involve more than two nodes 
(one sender and one or more recipients).
Depending on the type of transactional data, additional information on
each transaction may be available, such as a timestamps, message content,
recipient classes (e.g. To/Cc/Bcc) and other ``header'' information.

We focus on networks in which multiple transactions occur between nodes,
and each transaction (e.g. email) has a single sender and potentially multiple
recipients. Since email data is the most obvious application, we use
that language to develop our model. We shall assume a fixed number ($M$)
of nodes (people) in the network, and that each transaction involves at
least one recipient.  Additional transaction data (content, time-stamp,
etc.) will not be used.  Thus for a group of $M$ nodes, the observable
data takes the form of a list of transactions, with each transaction
having a sender and between $1$ and $M-1$ recipients.

Given a social network, two common tasks are discovering group
structure in the network and predicting future links between nodes.
Our model combines these two ideas, allowing transactions
between nodes to depend on group membership (the ``role'' played by sender
and receiver). Considering nodes in a social network, it is natural
to assume that each node can potentially play different roles while
interacting with different sets of nodes. It is also reasonable to assume
that the likelihood of an interaction between two nodes will depend on
the roles they have assumed at the time of communication.  One can see
that these two assumptions hold in many social networks. For example,
in a network constructed from emails exchanged in an academic settings,
it is easily observed that each person can choose multiple roles such
as professor, teaching assistant, research assistant, student, and
office staff.

We propose a hierarchical Bayesian block-model inspired by the mixed membership 
stochastic block-model~(MMSB)~\cite{AiroldiEtAl.2008} for transactional
network data (Transactional MMSB, or TMMSB).  Detailed explanation of the network structure is presented 
in section~\ref{sec:data_representation}. We develop our model in section~\ref{sec:model}. We
discuss inference, estimation and model choice 
in section~\ref{sec:inference}. We review the MMSB model and other related work in section~\ref{sec:background}. We then introduce a novel performance measure for soft clustering. 
Simulation results and results from two datasets are presented in
section~\ref{sec:analysis}. We conclude the paper with a summary of the model, scalability results and a discussion of future directions.

\section{Data and data representations}\label{sec:data_representation}
\begin{figure}[tbh]
  \begin{center}
\includegraphics[scale=0.8]{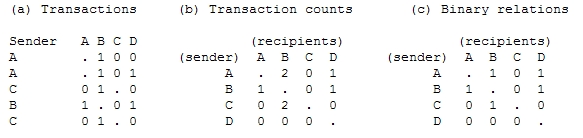} 
  \end{center}
  \caption{Simple example of transactional data, with various reductions and
representations.  (a) Original transactions (b) Matrix of transaction
counts (c) Binary relations obtained by thresholding counts at 1 (also
known as a {\em socio-matrix}.)  }
  \label{fig:ExampleData}
\end{figure}
In this section, we explain structure of the network data we seek to model. A toy 
example of such transactional data is represented in Figure~\ref{fig:ExampleData} (a).  We have $5$ 
transactions, each with a sender and one or more recipient.  We adopt the convention that the 
sender cannot be a recipient, and use a binary representation to identify recipients.  Thus the 
first message is from A to B (represented by a $1$ in the B column and $0$s in the C and D columns).  
The fourth message is from B to A and D. In general we shall assume $M$ nodes (here $M=4$) and $N$ transactions (here $N=5$).

\begin{figure}[tbh]
  \begin{center}
\includegraphics[scale=0.40]{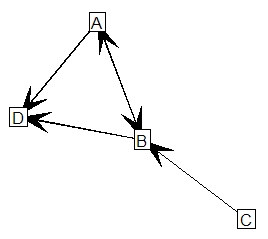}
  \end{center}
  \caption{Simple example: Network representation of socio-matrix of binary
relations from Figure~\ref{fig:ExampleData} (c).  
 } \label{fig:ExampleData2}
\end{figure}

Various summaries may be derived from the raw transactions, such as a matrix of 
transaction counts (number of messages for each sender/receiver pair), as in Figure~\ref{fig:ExampleData}(b). 
This could be converted into a matrix of binary relations by thresholding the number of messages 
(threshold of 1 used in Figure~\ref{fig:ExampleData}(c)).  This matrix of binary relations is often 
known as a {\em socio-matrix}. For networks with a small to moderate number of nodes, socio-matrices 
are visualized via a graph in which an edge indicates a directed relation between nodes. A simple visualization of the toy example is given in Figure~\ref{fig:ExampleData2}.

These summaries are ``lossy'' 
representations of transactional data. For instance, from the transaction counts, we know only that 
B received 2 messages from A, but not that D was a co-recipient of one of these messages. The 
socio-matrix loses additional information, since the counts are thresholded. 
One thing that is {\em not} lost by these representations is the directional nature of the relations.

Representations such as the frequency matrix and socio-matrix 
form the basis for some network models. For instance, the latent space approach of~\cite{HoffEtAl2002} 
and~\cite{HandcockEtAl2007} seeks a representation of nodes as points in a ``latent space'', with 
the probability of an edge between nodes as a decreasing function of the distance between their 
latent positions.  Extensions of the latent space model
for count data~\cite{KrivitskyHandcock2008} could be used on transaction counts.  The Mixed Membership Stochastic Block-model~\cite{AiroldiEtAl.2008} 
discussed in the later sections also seeks to model a binary socio-matrix.
\section{Transactional Mixed Membership Block-Model}\label{sec:model}
Observed network data are inherently variable, since transactions occur at random, and a finite 
sample of possible transactions are observed. Probabilistic generative models provide an efficient 
framework for modeling under uncertainty, by treating links as random and developing a probability 
model for their generation.

We develop a block-model for transactional network
data, using the language of email data.  We assume $N$ messages are sent
within a network of $M$ nodes.  Message $n$ has a sender $S_n$, and the
recipient list is represented by $M$ binary variables $Y_{n1}, \ldots,
Y_{nM}$, where $Y_{nm} =1$ indicates node $m$ received message $n$.
For each $n$, at least
one $Y_{nm}=1$, and if $S_n=i$, then $Y_{ni}=0$ (i.e. a sender doesn't
send to herself).  

Our model supposes the existence of $K$ groups. The probability of node $i$ sending a message to 
node $j$ is determined by the (unobserved) group memberships of sender $i$ and of receiver $j$.  
These probabilities are collected in a $K \times K$ {\em interaction matrix} $B$.  Element $B_{kl}$ 
is the probability of a node $i$ in group $k$ sending a message to a node $j$ in group $l$.
This defines a basic block-model as in~\cite{Wang.Wong.JASA.1987}:
probability of a relation is identical between all members of two groups. 

The ``mixed membership'' is incorporated into the model via an additional hierarchical level. 
Instead of assuming that each node belongs to just one group, the group membership of nodes is 
allowed to vary.  That is, the process for generating a transaction involves random selection of group memberships for each
node in the network:  A group label for the ``sender'' and a
group label for each potential ``recipient''.  Thus in 
a list of $N$ messages, node $i$ would have $N$
independent memberships sampled for it (one for each message). {\em Conditional} on these group memberships,
the $Y_{ij}$ are independent Bernoulli outcomes.  Node $i$ has a $K-$dimensional vector $\pi_i$
of membership probabilities for the $K$ classes, with $\sum_{k=1}^K
\pi_{ik}=1$.  The only observables for
this model are messages $Y_{ij}, i=1,\ldots,N, j=1, \ldots, M$ and senders $S_n$.
The matrix $B$ and group membership probabilities $\pi_1, \ldots, \pi_M$ 
must all be estimated.

Our generative model for transactional data is shown in Figure~\ref{EMMSB-Generative-Process}. Each 
node $i$ has a mixed membership vector $\pi_i$ which is drawn from a Dirichlet prior with hyperparameter 
$\alpha$. Generating a new email involves selecting a node to be the sender
from a multinomial distribution.
Although the ``friendship value'' 
mechanism  for selecting a sender is equivalent to a multinomial draw, we employ this more elaborate 
notation to enable subsequent generalization of the model.
 For each email $n$, each node $i$ samples 
its group $z_{ni}$ using its membership vector $\pi_i$.  We represent $z_{ni}$ as a binary $K$-dimensional 
vector with exactly one nonzero element.  The recipients of this email are sampled as $M-1$ Bernoulli 
random variables.   The Bernoulli probability $z_{nu} B z^T_{nj}$ indicates the selection of the element 
of $B$ corresponding to the current group membership of the sending node $u$ and the (potential) receiving 
node $j$. The group membership of a node mixes over time, however for each email, each node chooses to be a member of a single group.

\begin{figure}[tb]
\centering
\shadowbox{\begin{minipage}{\linewidth}
\begin{enumerate}
    \item For each node $i$, draw mixed-membership vector
        $\pi_i \sim \mbox{Dirichlet}(\alpha)$
    \item For each node $i$, draw its sender probability
$\lambda_i$.
    \item Choose $N \sim Poisson(\varepsilon)$: number of emails 
    \item For each email $n$
    \begin{enumerate}
	\item For each node $i$, draw $z_{ni} \sim \mbox{Multinomial}(\pi_i)$
       \item Pick node $u$ as sender (i.e., $S_n=u$) among all
the nodes with probability $\lambda_u$.
       \item For each node $j \neq u$,
          draw $Y_{n,j} \sim \mbox{Bernoulli}(z_{nu} B z^T_{nj})$
    \end{enumerate}
\end{enumerate}
\end{minipage}}
\caption{Generative process for Mixed Membership Stochastic Block Model
for Transactional Networks} \label{EMMSB-Generative-Process}
\end{figure}

The main input parameter of the model is the number of groups $K$. For a model with $K$ groups, other 
parameters that need to be estimated include the $K$-dimensional Dirichlet parameter $\alpha$ and a 
$K \times K$ interaction matrix $B$. Interaction matrix $B$ can be interpreted differently depending 
on the domain in which the model is applied. For email domain, $B_{kl}$ is the probability that a node 
from group $l$ will receive a message from a node in group $k$.
Since the $kl$ entry of the matrix $B$ corresponds to the probability
of a message being sent from a member of group $k$ to a member of group $l$, the only
restriction on $B$ is that entries must be between 0 and 1.  There are no
restrictions on rows, columns or other collections of $B$ elements.

The arbitrary form of $B$ allows the TMMSB model to
capture quite general forms of group behavior.
Possibilities for $B$ include:
\begin{enumerate}
\item Large diagonal elements, corresponding to groups that communicate
among themselves, but not with other groups.
\item Rows with some large entries, corresponding to groups defined by
high intensity of sending communication to specific other groups.
\item Columns with some large entries, corresponding to groups defined by
high intensity of receiving communication from specific other groups.
\item Small diagonal elements and some large off-diagonal elements,
corresponding to groups that do not communicate among themselves, and are
defined by similar communication patterns with members of some other
groups.
\end{enumerate}
Among other clustering models for socio-matrices, only the first notion of
clustering is common.  Section~\ref{sec:sims} illustrates some examples
of the structures for $B$ described above.

Combining the distributions specified in this section
gives a joint distribution over latent variables and the observations as

\begin{align*}
	p(Y, S, \pi_{1:M},  & \lambda_{1:M}, Z_{1:N, 1:M} | \alpha, B,\mu) = \\
	& \prod_{m=1}^M p(\pi_m | \alpha) \prod_{m=1}^M p(\lambda_m | \mu) 
	 \prod_{n=1}^N [  p(S_n | \lambda) \prod_{m=1}^M p(z_{nm}|
\pi_m)\times \\ 
	& \prod_{m=1, m \neq S_n}^M p( Y_{n,m} | Z_{nm}, Z_{nS_n}, B ) ]
\end{align*}
where $Z_{1:N, 1:M}$ is the set of group assignments for all nodes $1,\ldots,M$ in all messages $1,\ldots, N$. $Y$ 
is a $N\times M$ binary matrix in which every row is a transaction and ones in each row encode the 
recipients of the corresponding email.  Since our focus is estimating groups and membership, in the following
sections we condition on senders $S_n$, eliminating the need to infer the
$\lambda$'s.
\section{Inference and Model Choice}\label{sec:inference}
We derive empirical Bayes estimates for the $B$ parameter and use variational 
approximation inference. The posterior inference in our model is
intractable, nvolving a multidimensional integral and
summations:
\begin{align*}
p(Y|S, \alpha, B) = \int_{\pi} 	& \sum_{Z} \prod_{m=1}^M p(\pi_m | \alpha) \times \\
				& \prod_{n=1}^N [ \prod_{m=1}^M p(z_{nm}| \pi_m)  \prod_{m=1, m \neq S_n}^M p( Y_{n,m} | Z_{nm}, Z_{nS_n}, B ) ] d\pi
\end{align*}

For using variational methods for inference, we pick a distribution over
latent variables with free parameters. This distribution which is often
called the variational distribution then approximates the true posterior in
terms of Kullback-Leibler divergence by fitting its free parameters. We use
a fully-factorized mean-field family of distributions as our variational
distribution:
\begin{align}
	q(\pi_{1:M}, Z_{1:N,1:M}) = & \prod_{m=1}^M q_1(\pi_m | \gamma_m) 
\prod_{n=1}^N \prod_{m=1}^M q_2(z_{n,m} | \phi_{n,m}) 
 \label{eq:variational} 
	& 
\end{align}
where $q_1$ is a Dirichlet, 
 and $q_2$ is a Multinomial distribution.
$\{\gamma_{1:M}, \phi_{1:N,1:M}\}$ is the set of variational parameters that will 
be optimized to tighten the bound between the true posterior and variational distribution. 

The updates for variational parameters $\phi_{nm}$ and $\gamma_m$ are
\begin{align}
	\phi_{nm,k} &\propto \mathbbm{E}_q(\log(\pi_{m,k})) \times \\
	& \indicator{m \neq S_n} . \prod_{l=1}^K \left(  B_{lk}^{Y_{nm}}.(1-B_{lk})^{1-Y_{nm}} \right) ^{\phi_{nS_n,l}} \times \nonumber \\
	& \indicator{m = S_n} . \prod_{m^\prime \neq m} \prod_{l=1}^K \left(  B_{kl}^{Y_{nm^\prime}}.(1-B_{kl})^{1-Y_{nm^\prime}} \right) ^{\phi_{nm^\prime,l}} \nonumber
\end{align}
for all transactions $n=1, \ldots, N$ and all nodes $m=1,\ldots, M$, and
\begin{equation}\label{eq:gamma}
\gamma_{m,k} = \alpha_k + \sum_{n=1}^N \phi_{nm,k}
\end{equation}
for all nodes $m=1,\ldots, M$. 
The empirical Bayes estimate for $B$ parameter is
\begin{equation}
	B_{k,l} = \frac{\sum_{n=1}^N \sum_{m=1, m \neq S_n}^M \phi_{nS_n,k} \phi_{nm,l} Y_{nm}}{ \sum_{n=1}^N \sum_{m=1, m \neq S_n}^M \phi_{nS_n,k} \phi_{nm,l} }
\label{eq:EBB}
\end{equation}
Inference for multinomial sender probability $\lambda$ is straightforward and thus
omitted.
We fix $\alpha=0.1$ in our inference.

Algorithm~\ref{TMMSB_VB_Algorithm}  shows the pseudocode for the
variational EM inference for the proposed model. For simplicity, a stylized
version of the algorithm is presented. 

\begin{algorithm}[tbhp]
\caption{VB Inference Algorithm} \label{TMMSB_VB_Algorithm}
\begin{algorithmic}
{\footnotesize 
\STATE Initialize $\gamma_{mk} = N/K$ for all $m=1,\ldots,M$ and $k=1,\ldots,K$\;
\STATE Initialize $\phi_{nmk} = 1/K$ for all $n=1,\ldots,N$, $m=1,\ldots,M$ and $k=1,\ldots,K$\;
\COMMENT{$\phi$ can be initialized using the output of clustering method mentioned in Section\ref{sec:RedditLinkPrediction}}
\STATE Fix $\alpha=0.1$
\REPEAT{
  \STATE Estimate $B$ matrix using Eq.~\ref{eq:EBB}

  \REPEAT{
    \FOR{$n \leftarrow 1$ to $N$}
      \FOR{$m \leftarrow 1$ to $M$}
	\FOR{$k \leftarrow 1$ to $K$}
	  \STATE Estimate $\phi_{nm,k}$ using Eq.~\ref{eq:Phi}
	\ENDFOR
	\STATE Normalize $\phi_{nm,k}$ for $k=1,\ldots,K$ to sum to $1$
      \ENDFOR
    \ENDFOR
    \FOR{$m \leftarrow 1$ to $M$}
      \FOR{$k \leftarrow 1$ to $K$}
	\STATE Estimate $\gamma_{m,k}$ using Eq.~\ref{eq:gamma}\
      \ENDFOR
    \ENDFOR
  } \UNTIL{convergence or a maximum number of iterations is reached}
} \UNTIL{convergence or a maximum number of iterations is reached}

\COMMENT{Convergence is reached when the change in likelihood $\leq$ some threshold}
}
\end{algorithmic}
\end{algorithm}

The inference algorithm described above is for a fixed
number of clusters $K$.  In order to choose the number of clusters, we
develop a BIC criterion, composed of a log-likelihood and a penalty term.

The log-likelihood of the model is a sum of two
terms, a ``sending'' term corresponding to selection of the sender node for each
transaction and a ``receiving''
term for choosing group memberships and which of the
$M-1$ other nodes receive the email.  We focus on the ``receiving'' term.
Conditional on the sender of a particular message, the likelihood for recipient nodes is
equivalent to $M-1$ Bernoulli trials to decide whether each node receives
the email (excluding the sender).  Since the memberships are unobserved, we
calculate a receiving probability as an average over group memberships.
That is, we compute the predicted
probabilities of $\Pr(j \mbox{ receives} | i \mbox{ sends}) = p_{ij} =
\pi_i B \pi_j^T$, for each node $i$ as a sender and all the other nodes
$j$. Then, we can write the ``receiving'' term of the likelihood as
\begin{equation}
	{\cal L} = \prod_{n=1}^N \prod_{j \in 1..M, j \neq S_n}
p_{ij}^{Y_{nj}}(1-p_{ij})^{1-Y_{nj}}
\end{equation}
where $S_n$ is the sender node for transaction $n$. Based on this predictive likelihood, we use the following approximation for the BIC score for choosing the number of groups:
\begin{equation}
	BIC = 2.\log {\cal L} - ( K^2 + K).\log(|Y|),
\end{equation}
where $ K^2 + K$ is the number of parameters in the model (elements of $B$
and $\alpha$) and $|Y|=\sum_{n,m} y_{n,m}$ is the number of total
recipients in the network.
\section{Related Research}\label{sec:background}
Our proposed model is inspired by the Mixed Membership Stochastic Block-model~(MMSB)~\cite{AiroldiEtAl.2008}. 
The MMSB model describes directional
binary-valued relations between sender/receiver pairs of nodes. It seeks to model socio-matrices, such as panel (c) of
Figure~\ref{fig:ExampleData}. For every sender/receiver pair, a single binary relation $w_{ij}$ is observed. 
If $w_{ij}=1$, a $i\rightarrow j$ relation has been observed; $w_{ij}=0$
indicates no relation.  The $w_{ij}$ are modeled as conditionally
independent Bernoulli outcomes, with $\Pr(w_{ij}=1) = p_{ij}$.  
Mixed membership behaviour is incorporated by allowing
node membership of nodes to change every time a directed relation is
sampled.  A matrix similar to $B$ represents edge probabilities between
nodes in a directed binary relation.

Direct application of the MMSB model to transactional data would require 
simplification of the raw data.  For instance, in~\cite{Huh.Fienberg.2008},
directional binary-valued $i\rightarrow j$ relations are generated between
pairs of nodes by counting the number of messages sent by $i$ and
received by $j$, and thresholding these counts at a specified level. This corresponds to the simplification from Figure~\ref{fig:ExampleData}(a)
to (c). This simplification discards 
co-recipient information and weakens message frequency information.

Other papers have studied modeling of transactional
data. Prediction of link strength in a Facebook network is studied
in \cite{Kahanda.Neville.ICWSM.2009}.  In their comparative study,
transactional data on a network are used as features in prediction
of a binary "top friend" relation.  Specific models for prediction of
transactions are not developed. There has been recent work considering
frequency of interactions for modeling. In~\cite{kurihara.2006}, a
stochastic block model is proposed for pairwise  relation networks in
which the frequency of relations are taken into account.  The number of
groups is inferred using Dirichlet process priors.  Multiple recipient
transactions are not considered in \cite{kurihara.2006}.
\section{Novel Clustering Performance Measures}\label{sec:Measures}
In order to assess clustering performance of our model, some new measures
are needed.  We consider a situation in which ``ground truth'' is available
in the data, in the form of (possibly soft) class labels for each node.
Thus measures that can assess the similarity of two different soft labels
(one from a model and one from data) are needed.  Although these are
developed in the context of our model, these new measures can compare
any two soft clusterings.  These measures will be applied later in
Section~\ref{sec:RedditClust}.

We wish to compare a predicted soft clustering (mixed
membership vector $\pi_i$ in our model) with an
observed mixed membership vector (e.g. normalized frequencies in the
reddit case; see Section~\ref{sec:Reddit}).
We propose an novel extension to the evaluation measures developed
in~\cite{Amigo.Gonzalo.Artiles.Artiles:2009}. Their evaluation measure
is expressed in terms of precision, recall and F-Measure values developed for 
overlapping clustering output.  By ``overlapping'', we mean a 0/1
assignment in which nodes can be assigned to multiple
clusters. \textcolor{black}{Our extended set of measures can be used to compare ``soft'' clustering results.}

The proposed metrics in~\cite{Amigo.Gonzalo.Artiles.Artiles:2009} for overlapping clustering are extensions 
of the BCubed metrics~\cite{BaggaBaldwin1998}. BCubed metrics measure precision and recall for each data point. The precision 
of a data point $i$ is the fraction of data points assigned to the same cluster as $i$ which belong 
to the same true class as $i$. Recall for $i$ is the fraction of data points from the same true class 
that are assigned to the same cluster as $i$. Extensions of precision and recall for overlapping clustering 
are defined as follows:
$$
  Precision(e,e^\prime) = \frac{Min( |C(e) \cap C(e^\prime)|, |L(e) \cap
L(e^\prime)| )}{|C(e) \cap C(e^\prime)|}, $$
  $$ Recall(e,e^\prime)    = \frac{Min( |C(e) \cap C(e^\prime)|, |L(e) \cap
L(e^\prime)| )}{|L(e) \cap L(e^\prime)|}, $$
where $e$ and $e^\prime$ are two data points, $L(e)$ is the set of classes and $C(e)$ is the set of 
clusters assigned to $e$.  The expression $|C(e) \cap C(e^\prime)|$ counts
the number of classes common to $e$ and $e^\prime$.
In our case, the points are not assigned to multiple clusters or do not belong 
to multiple classes. Each point has a membership probability vector assigned to it by the model and 
it has a true membership probability vector. We extend the metrics above to this case as follows:
  $$Precision(e,e^\prime) = \frac{Min( \pi(e).\pi(e^\prime),
\gamma(e).\gamma(e^\prime) )}{\pi(e).\pi(e^\prime)}  $$
  $$Recall(e,e^\prime)    = \frac{Min( \pi(e).\pi(e^\prime),
\gamma(e).\gamma(e^\prime) )}{\gamma(e).\gamma(e^\prime)} $$
where $\pi(e)$ is the estimated membership probability vector,
$\gamma(e)$ is the true membership 
probability vector for data point $e$, and $a.b = a^Tb$ for two vectors $a$
and $b$.  Aggregate precision and recall measures are obtained by averaging
over all pairs of nodes.  F-measure is defined as the harmonic mean of precision and recall.

\section{Examples}\label{sec:analysis}
In this section, we present experimental results on simulated networks, 
an email network based on the Enron corpus and another transactional network corpus built 
from the social news website \url{www.reddit.com}. 
\subsection{Simulation Results}\label{sec:sims}
We simulate four transactional networks, and verify that the learned
models recover the true model parameters. \textcolor{black}{We are
particularly interested in two parameters, membership probabilities
of nodes ($\pi_i$'s) and the $B$ interaction matrix.} The simulation parameters
are listed in Table~\ref{tb:simulated_data}. For $\alpha=0.05$,
node membership probabilities are concentrated in one group.
When $\alpha=0.25$, many nodes display mixed membership.

\begin{table}[ht]
	\centering
	\begin{tabular}{r|rrrr}
	Dataset	& $K$	& $\alpha$	& $M$	& $N$ \\
	\hline
	$1$		& $3$	& $0.05$	& $50$	& $500$ \\
	$2$		& $4$	& $0.05$	& $65$	& $650$ \\
	$3$		& $4$	& $0.25$	& $65$	& $650$ \\
	$4$		& $9$	& $0.05$	& $150$	& $1500$ \\		
	\end{tabular}
	\caption{Simulated datasets. Each network contains $M$ nodes and
has $N$ transactions.}\label{tb:simulated_data} 
\end{table}
In all four cases, the recovered $B$ matrix is very close to the
actual matrix used for simulation. We focus here on results for $K=4,
\alpha=0.25$.  This is the most challenging scenario, since most
nodes have mixed membership. The true and recovered $B$ matrices
(Table~\ref{tb:TrueEstimatedB}) are very close.  In this and the other
simulations large off-diagonal entries are present in $B$, implying
that some groups are  defined by high volume of communication to nodes
belonging to {\em other} groups.

\begin{table*}[ht]
	\begin{minipage}[b]{0.45\linewidth}\centering
		\begin{tabular}{|rrrr|}
		\hline
		0.01  & \textbf{0.2}   & 0.01  & 0.01 \\
		0.01  & \textbf{0.3}   & \textbf{0.2}   & \textbf{0.1} \\
		\textbf{0.1}   & 0.01  & 0.01  & \textbf{0.3} \\
		\textbf{0.1}   & 0.01  & 0.01  & \textbf{0.3} \\
		\hline
		\end{tabular}
	\end{minipage}
	\hspace{0.5cm}
	\begin{minipage}[b]{0.45\linewidth}
	\centering
		\begin{tabular}{|rrrr|}
		\hline
		0.0127  & \textbf{0.2012}   & 0.0149  & 0.0115 \\
		0.0064  & \textbf{0.3055}   & \textbf{0.2064}   & \textbf{0.0802} \\
		\textbf{0.0964}   & 0.0207  & 0.0146  & \textbf{0.2959} \\
		\textbf{0.0979}   & 0.0243  & 0.0164  & \textbf{0.2733} \\
		\hline
		\end{tabular}
	\end{minipage}
	\caption{Simulated data.  Left: $B$ matrix used to simulate the
network. Right: Estimated $B$ matrix from our inference algorithm}\label{tb:TrueEstimatedB}
\end{table*}
\begin{table*}[tbh]
	\centering
	\begin{tabular}{r|rrrrrrrrrr}
	$K^*$					& $2$		& $3$		& 		$4$	& 		$5$	& 		$6$	& 		$7$	\\
	\hline
BIC ($\times 10^4$)	& $-3.0357$	& $-2.9660$	& $-2.9229$	&
$-2.9229$	& $-2.9339$	& $-2.9501$	\\

	\end{tabular}
	\caption{BIC scores for different number of groups $K^*$ on the simulated dataset ($k=4$ and $\alpha=0.25$)}\label{tb:SimulationBIC}
\end{table*}

We also report the BIC scores for data simulated in the case $K=4$ and
$\alpha=0.25$. In this case, we know the actual number of groups but many
nodes have mixed membership. Therefore, predicting the number of groups is
more challenging compared to other cases where the group memberships are
close to certain. We estimate the parameters of the model using the
simulated data assuming $K^*=2, 3, \ldots, 7$ groups.
Table~\ref{tb:SimulationBIC} shows the BIC scores. The largest BIC values
correspond to $K=4$ (the actual value used for simulation) and $K=5$.

\begin{figure*}[htp]
  \begin{center}
    \subfigure[$K=3, \alpha = 0.05$]{\label{fig:simulation-adj-mat-a}\includegraphics[scale=0.15]{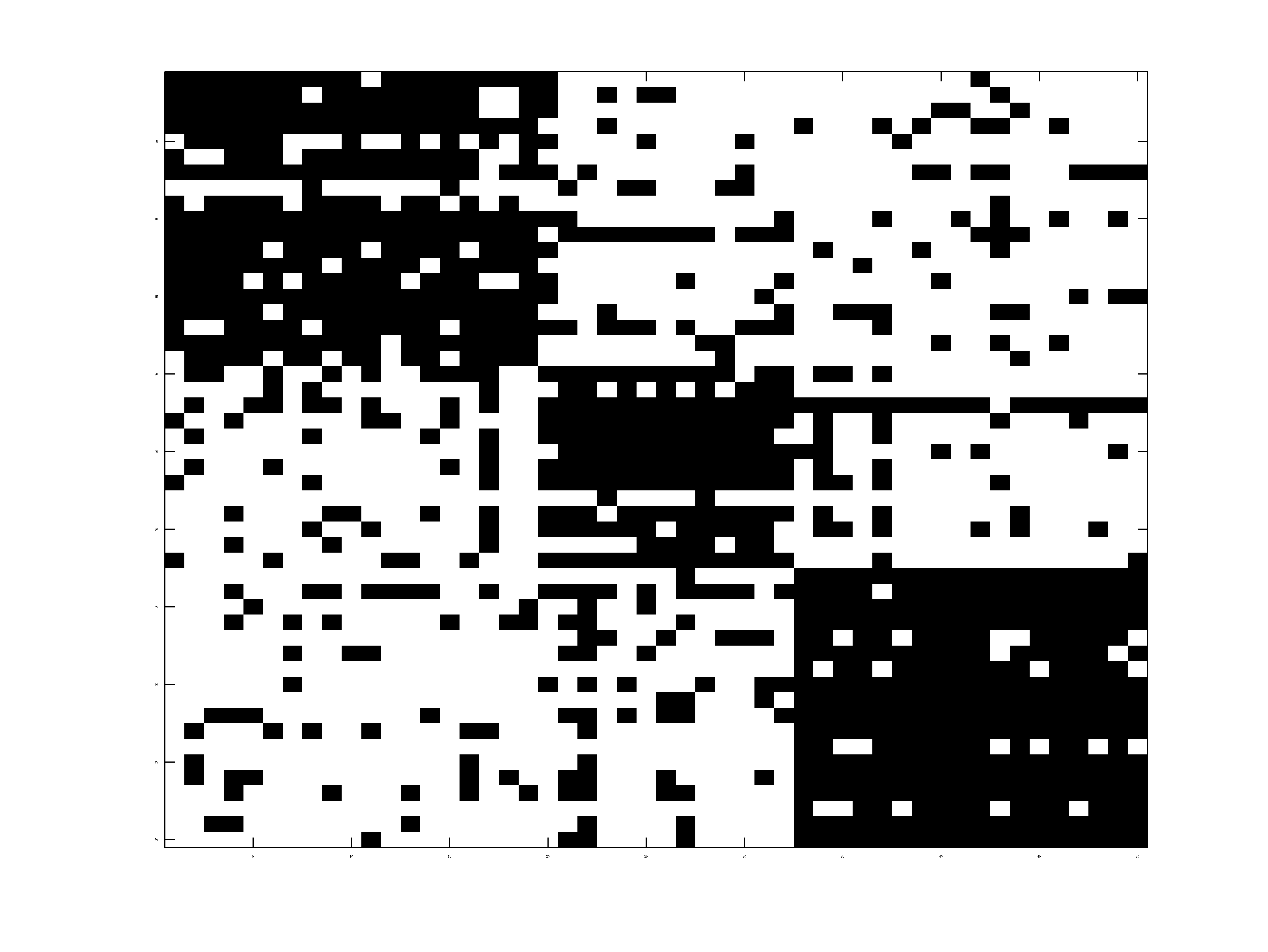}}  
    \subfigure[$K=4, \alpha = 0.05$]{\label{fig:simulation-adj-mat-b}\includegraphics[scale=0.15]{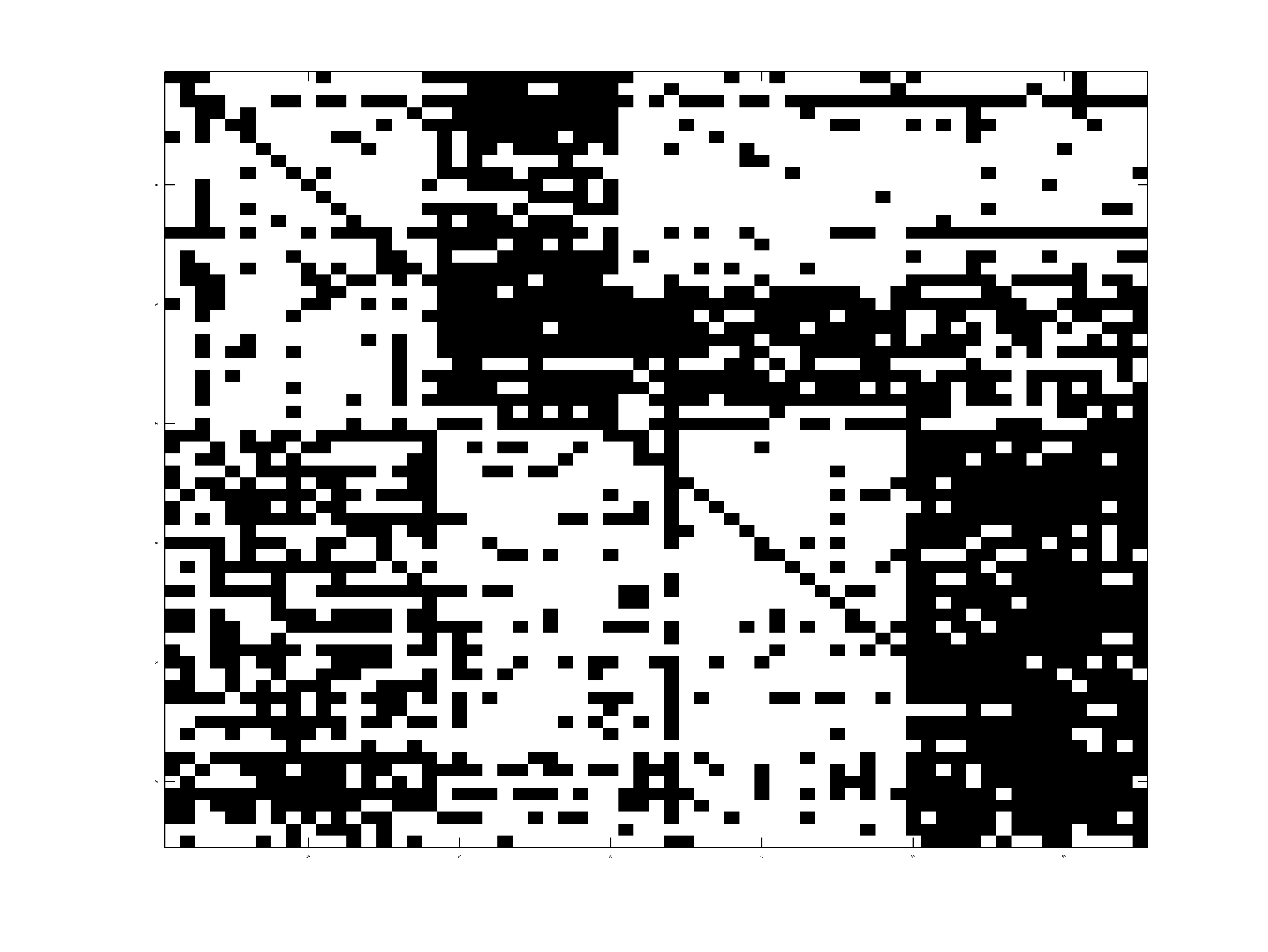}}
    \subfigure[$K=4, \alpha = 0.25$]{\label{fig:simulation-adj-mat-c}\includegraphics[scale=0.15]{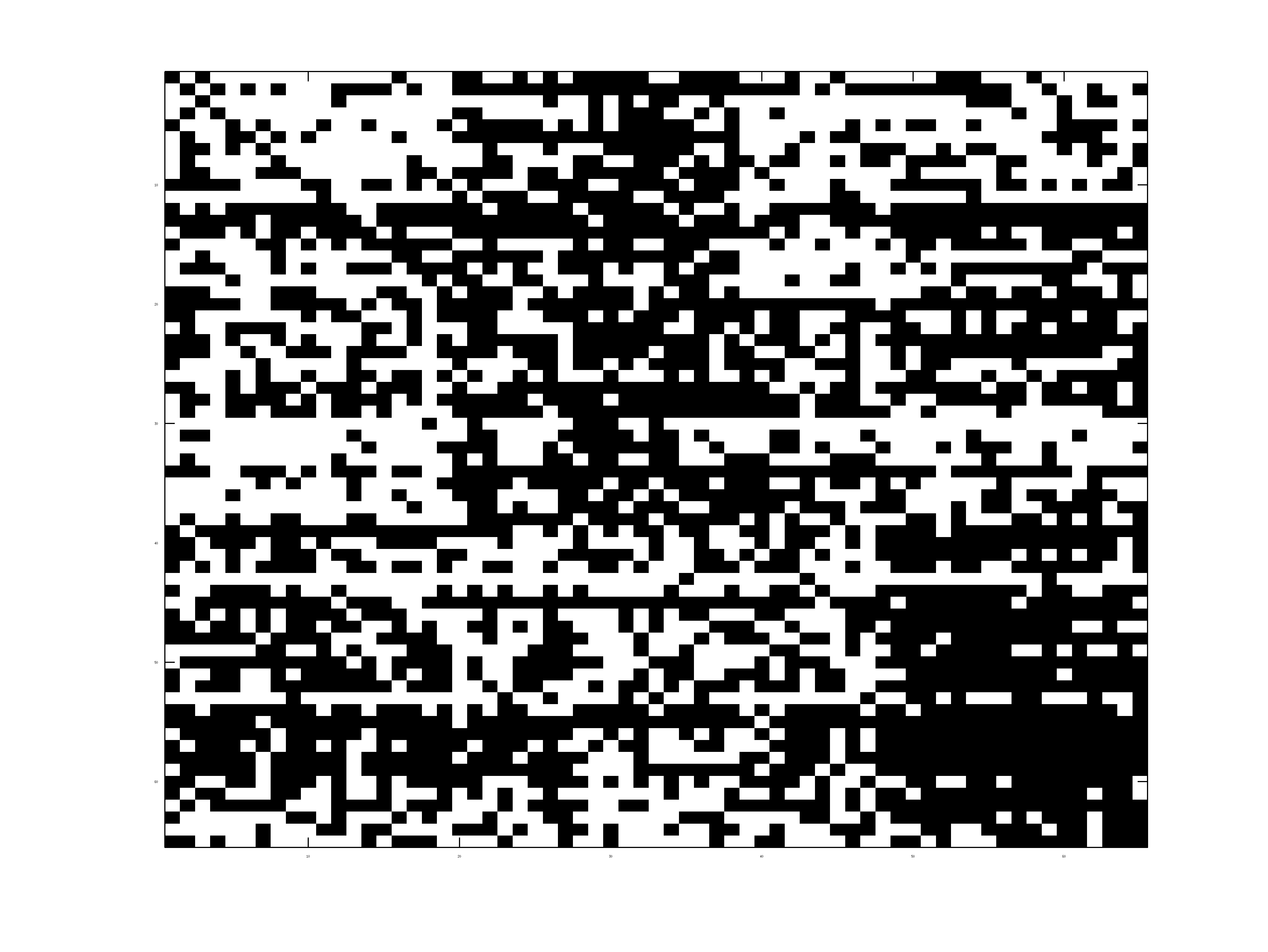}} 
    \subfigure[$K=9, \alpha = 0.05$]{\label{fig:simulation-adj-mat-d}\includegraphics[scale=0.15]{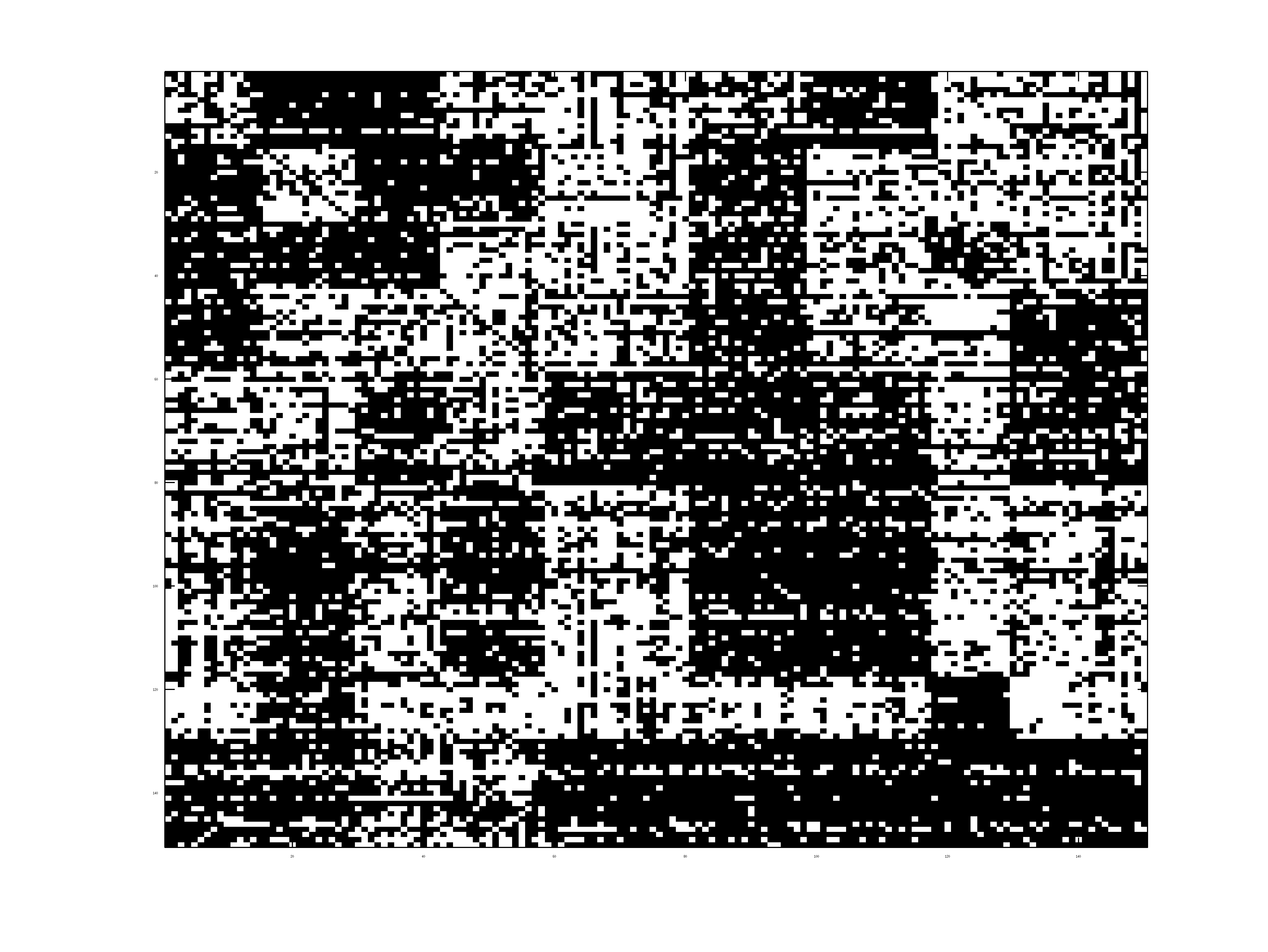}} \\
    \subfigure[$K=3, \alpha = 0.05$]{\label{fig:simulation-adj-mat-e}\includegraphics[scale=0.15]{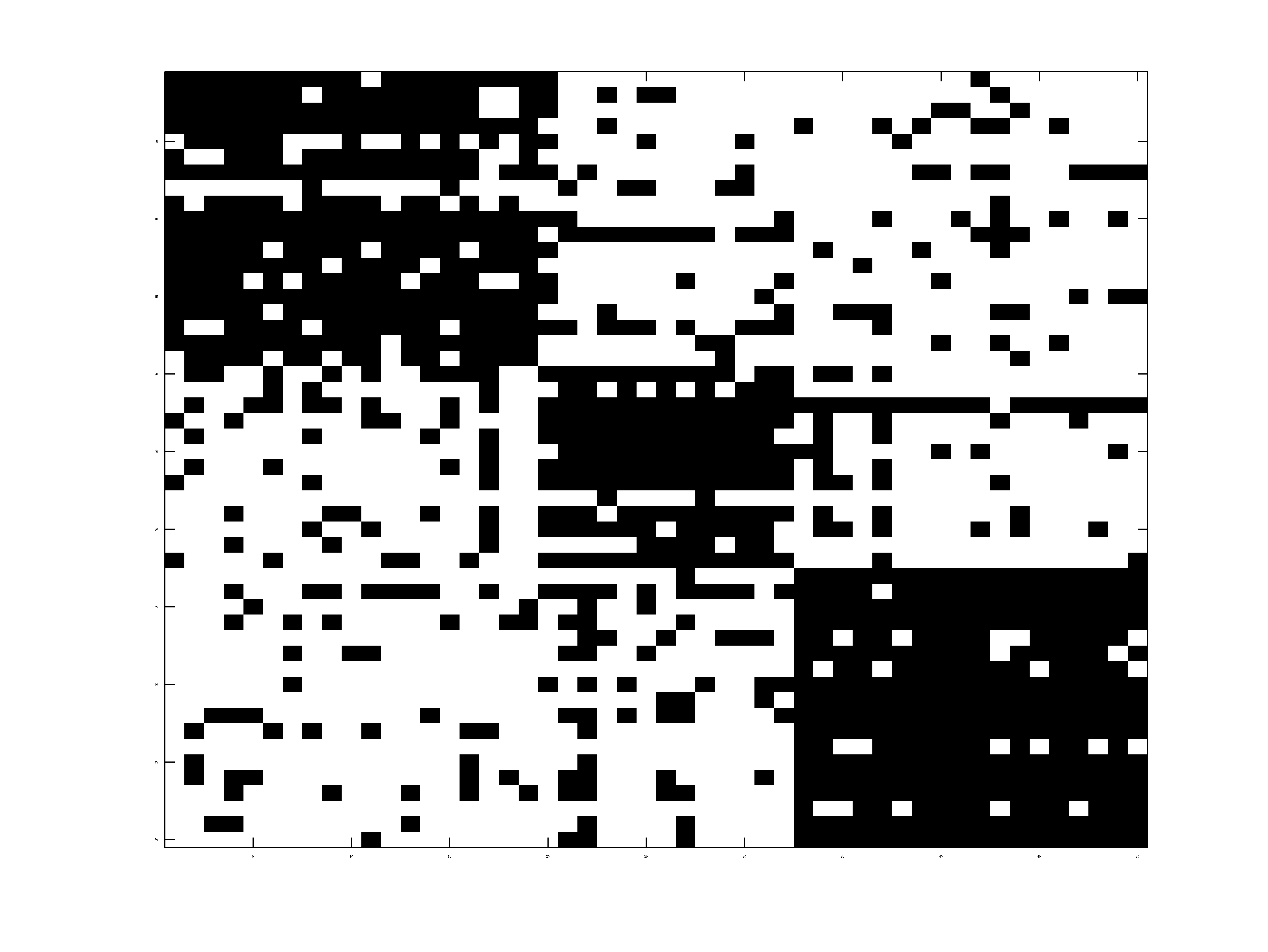}} 
    \subfigure[$K=4, \alpha = 0.05$]{\label{fig:simulation-adj-mat-f}\includegraphics[scale=0.15]{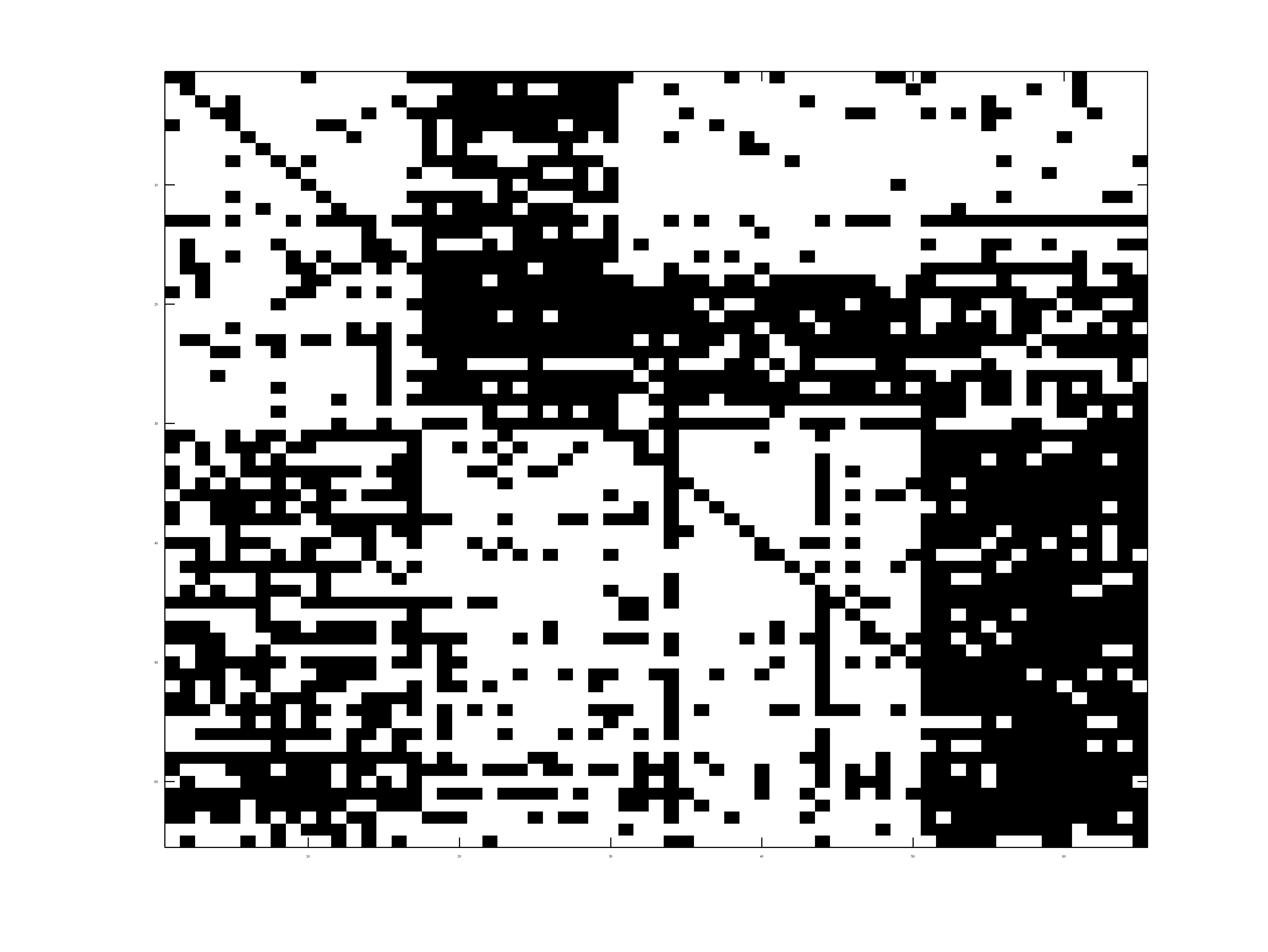}}
    \subfigure[$K=4, \alpha = 0.25$]{\label{fig:simulation-adj-mat-g}\includegraphics[scale=0.15]{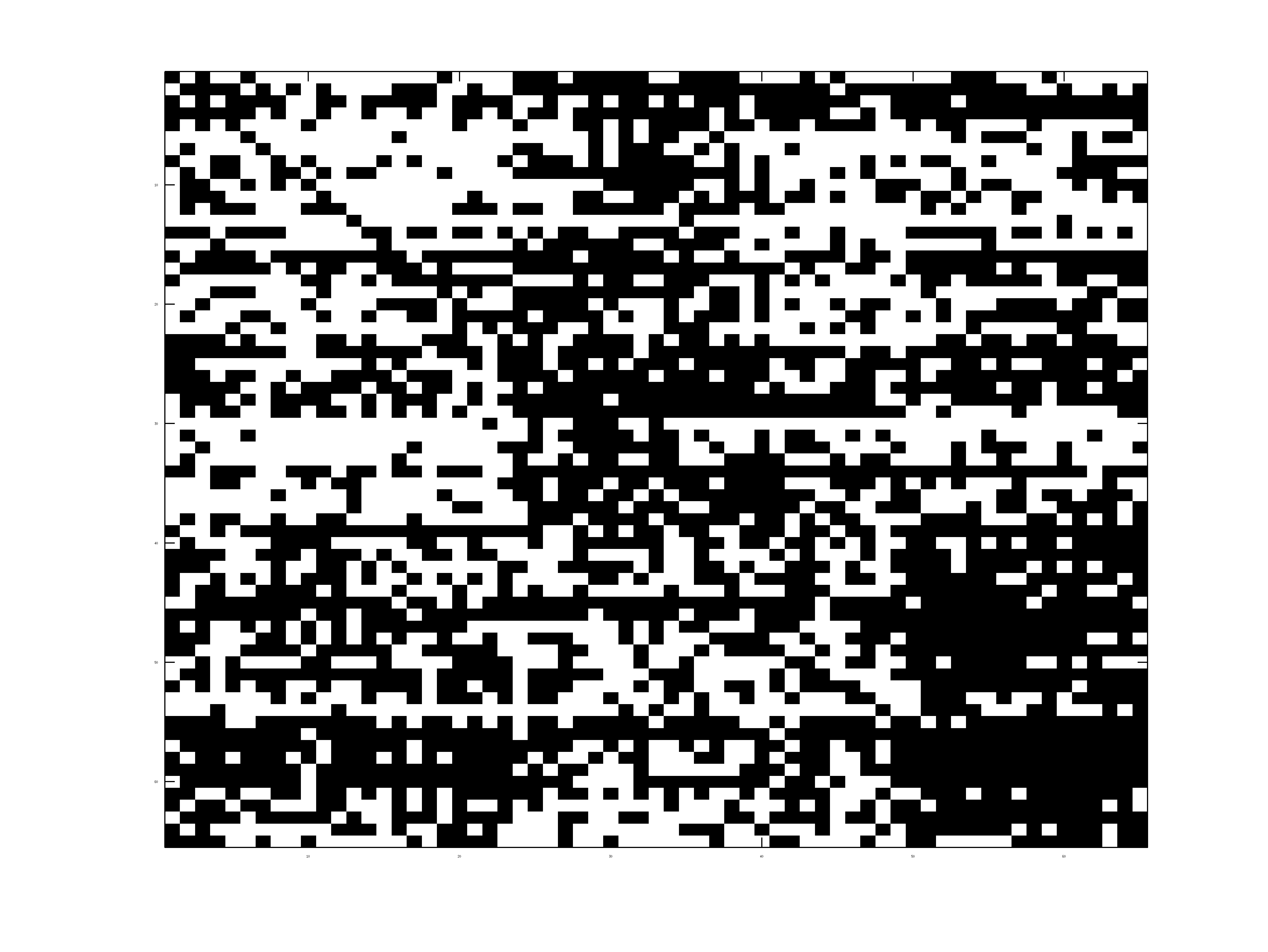}} 
    \subfigure[$K=9, \alpha = 0.05$]{\label{fig:simulation-adj-mat-h}\includegraphics[scale=0.15]{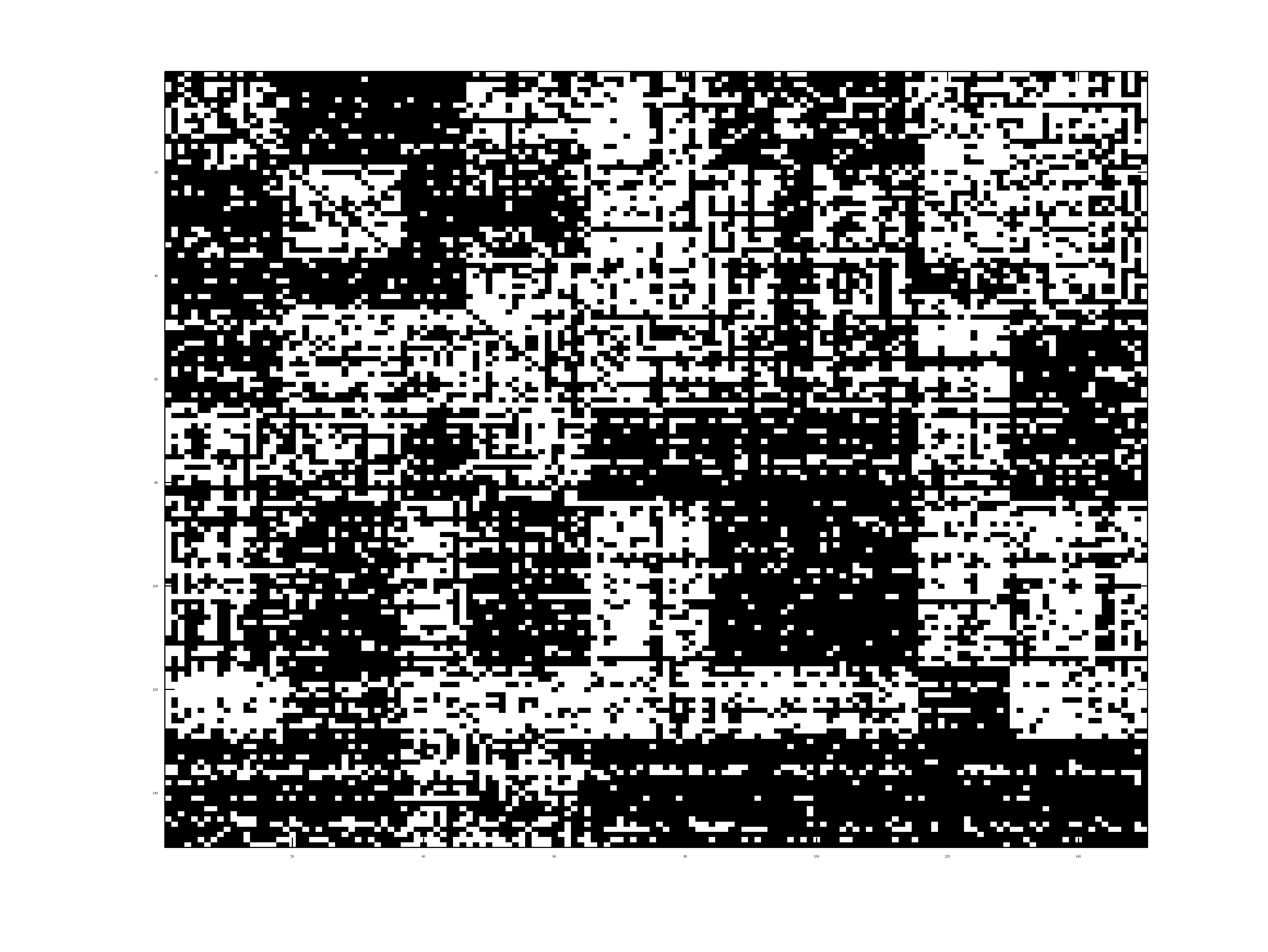}}     
  \end{center}
  \caption{Adjacency matrices, four simulated examples.  White cells have 0
messages, black have 1 or more.  Each $N\times N$ matrix is arranged
according to groups suggested by true $\pi$ (plots (a) - (d)), and estimates
$\hat{\pi}$ (plots (e) - (h)).
}
  \label{fig:simulation-adj-mat}
\end{figure*}

To assess the learned models, we use the estimated $\pi$ vectors to arrange
the data according to the most probable grouping of nodes.  The first row of Figure~\ref{fig:simulation-adj-mat} shows
the adjacency matrix for the simulated networks mentioned in
Table~\ref{tb:simulated_data}.  The $ij$ element of the adjacency matrix
is 1 if 1 or more message from node $i$ is received by node  $j$, and 0
otherwise.  Nodes are ordered along rows and columns according to their
most likely membership, as determined by {\em true} values of $\pi_i$.
The second row of the figure shows the same adjacency matrices, with
rows and columns ordered according to estimates $\hat{\pi}_{i}$. Similarity between top/bottom pairs in the figure indicates
that the inference algorithm is capable of recovering node memberships.

\begin{figure*}[tbh]
  \begin{center}
    \subfigure[$\pi_1$]{\label{fig:simulation-gamma-K4-a}\includegraphics[scale=0.15]{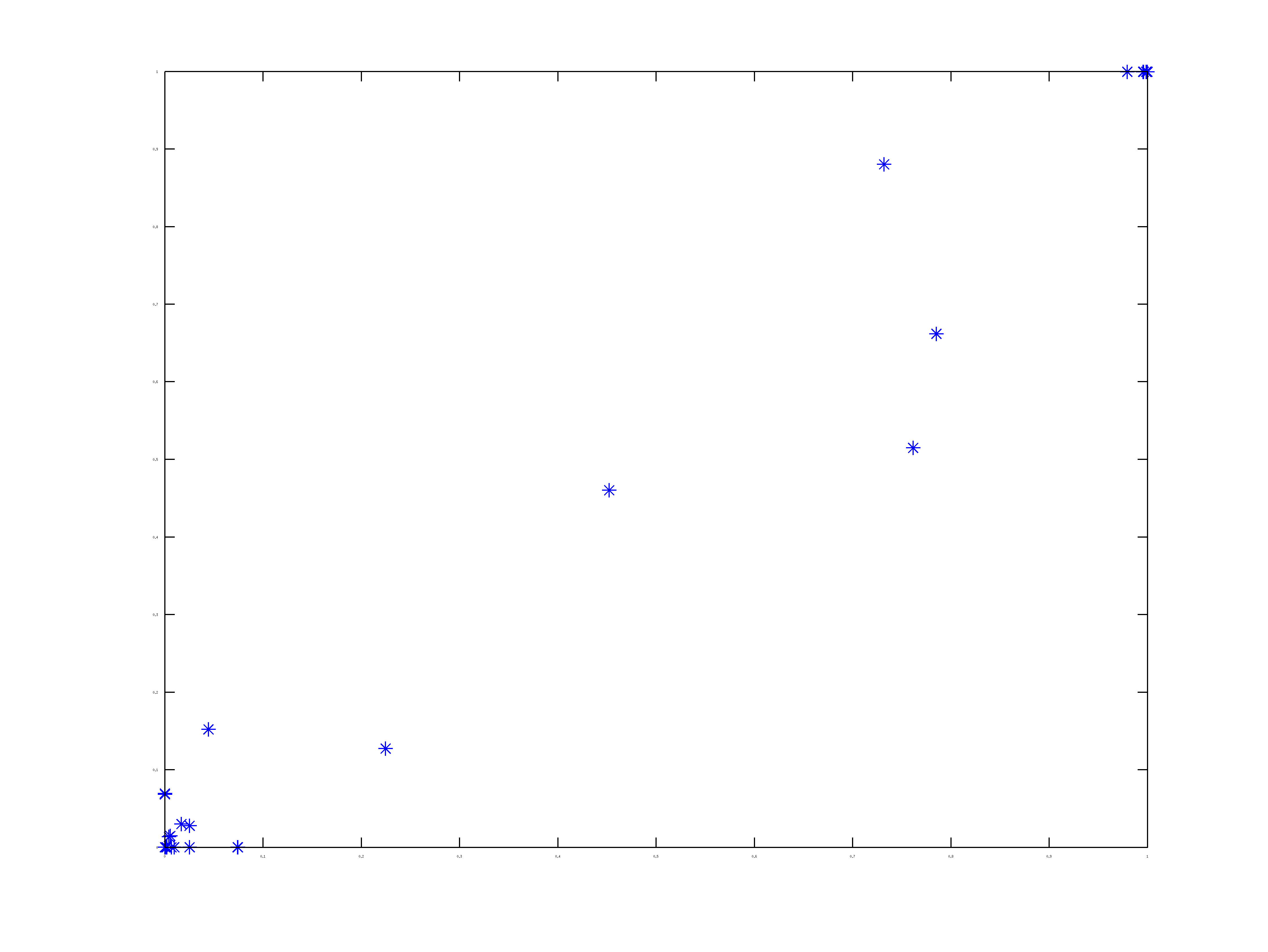}}  
    \subfigure[$\pi_2$]{\label{fig:simulation-gamma-K4-b}\includegraphics[scale=0.15]{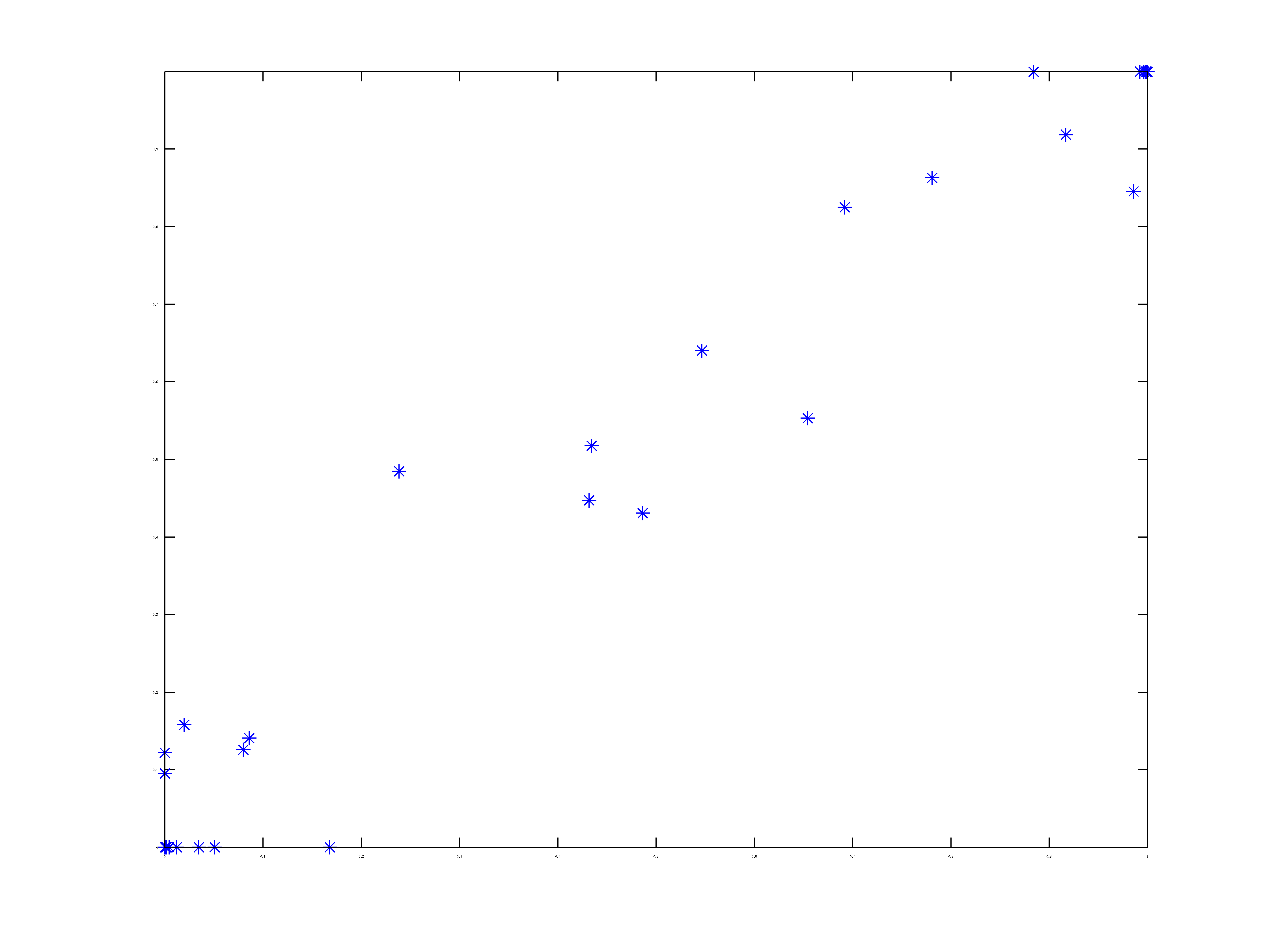}}
    \subfigure[$\pi_3$]{\label{fig:simulation-gamma-K4-c}\includegraphics[scale=0.15]{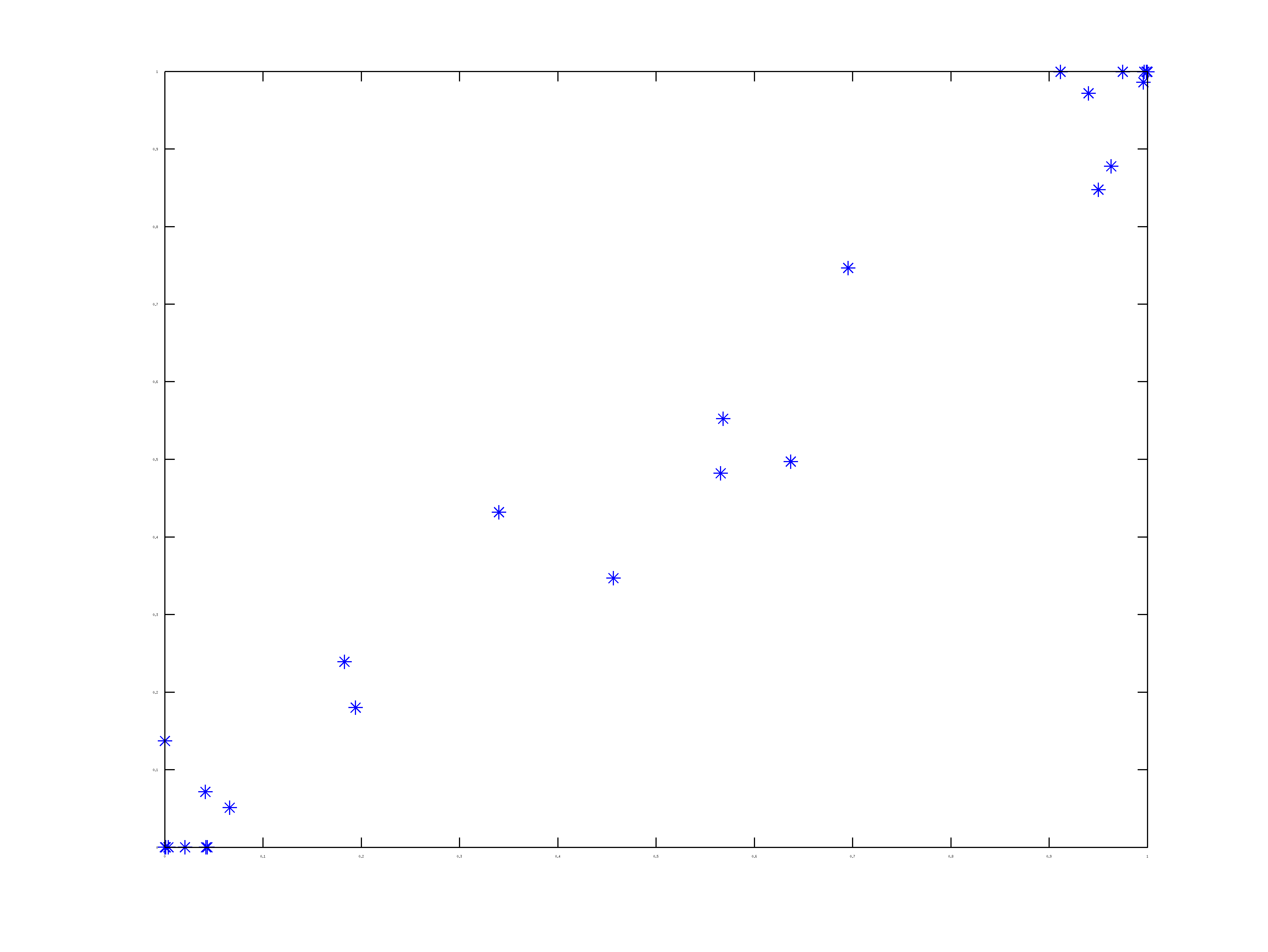}} 
    \subfigure[$\pi_4$]{\label{fig:simulation-gamma-K4-d}\includegraphics[scale=0.15]{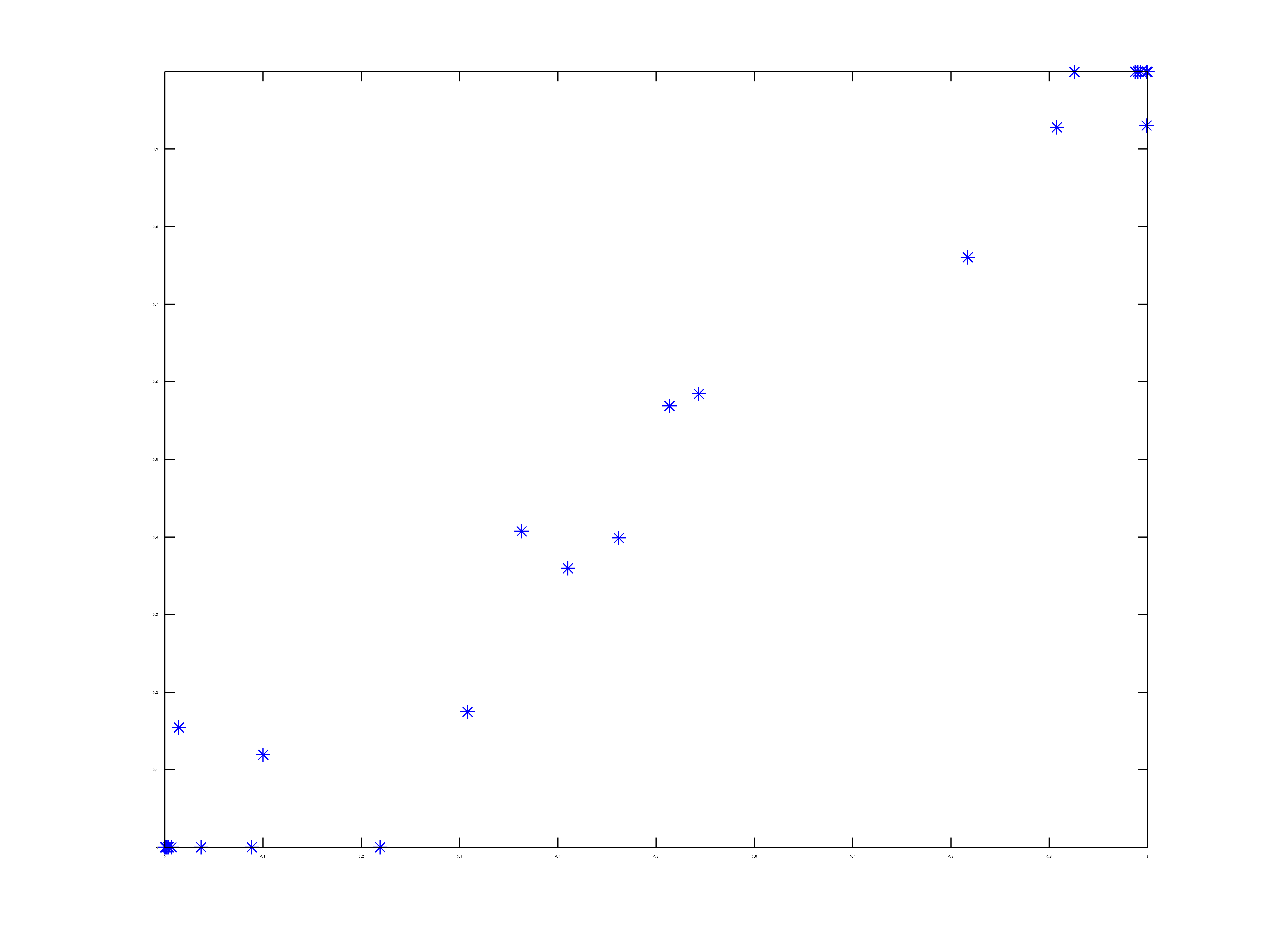}} \\
    \subfigure[$\pi_1$]{\label{fig:simulation-gamma-K4-e}\includegraphics[scale=0.15]{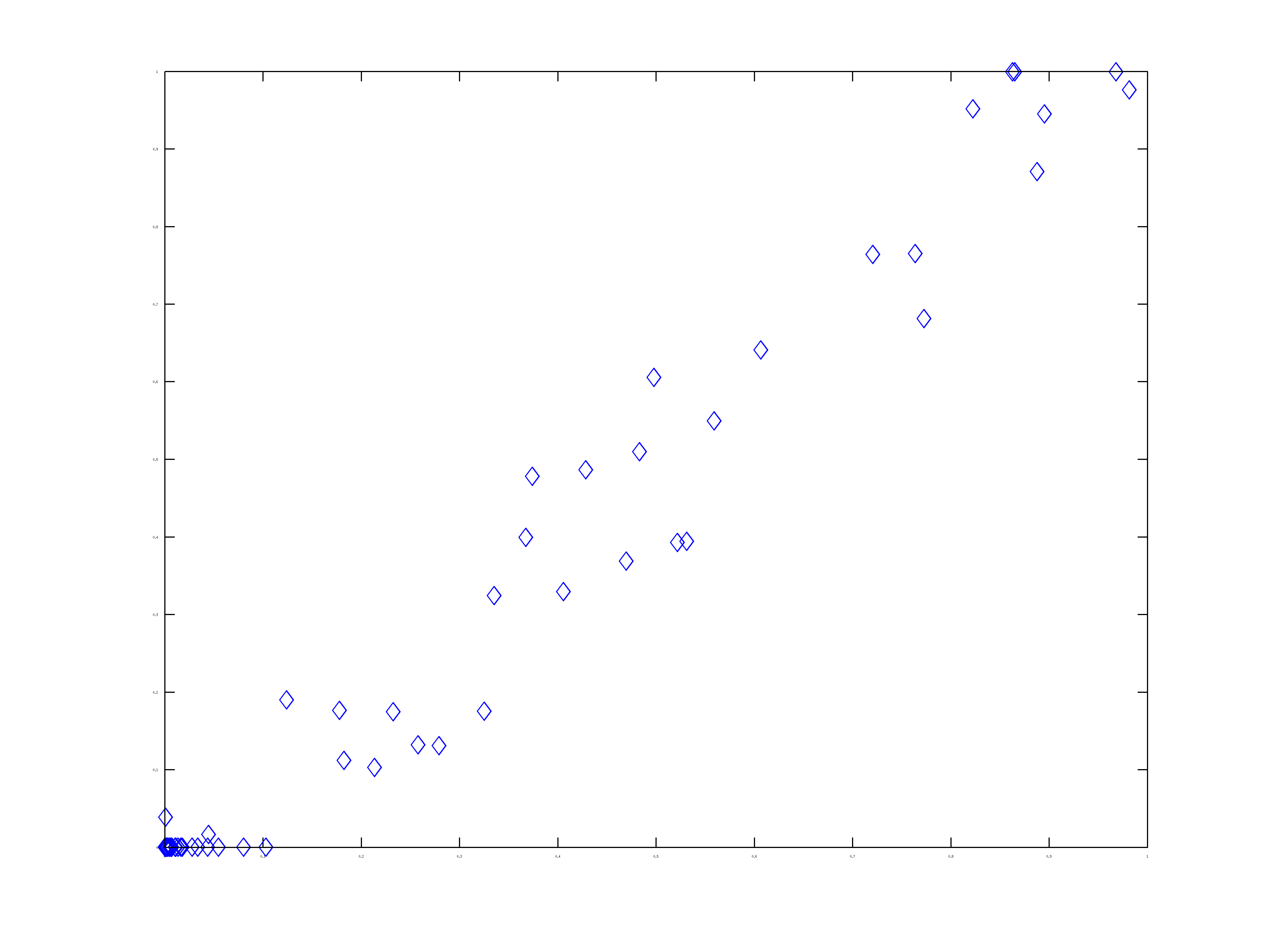}}    
    \subfigure[$\pi_2$]{\label{fig:simulation-gamma-K4-f}\includegraphics[scale=0.15]{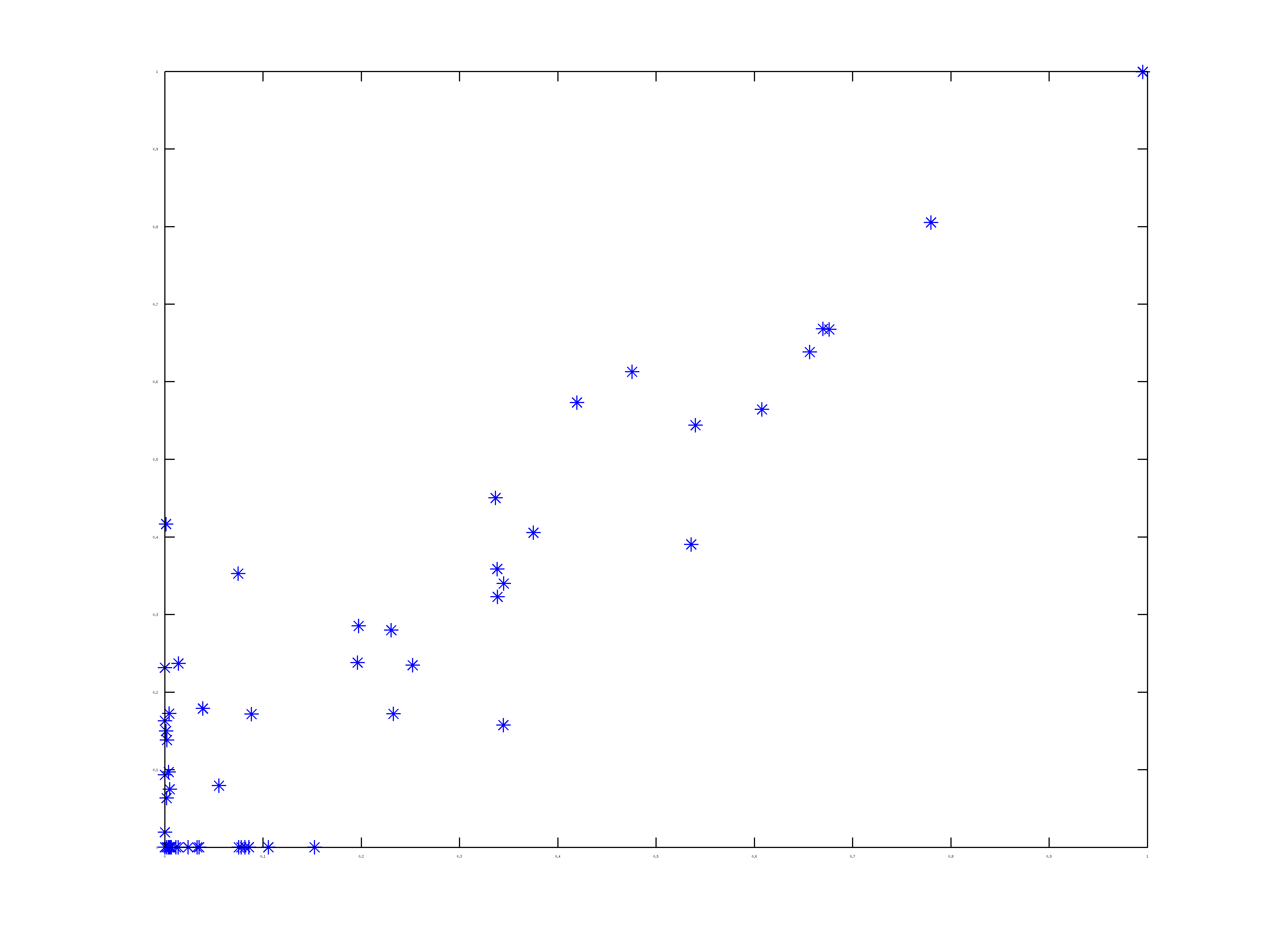}}
    \subfigure[$\pi_3$]{\label{fig:simulation-gamma-K4-g}\includegraphics[scale=0.15]{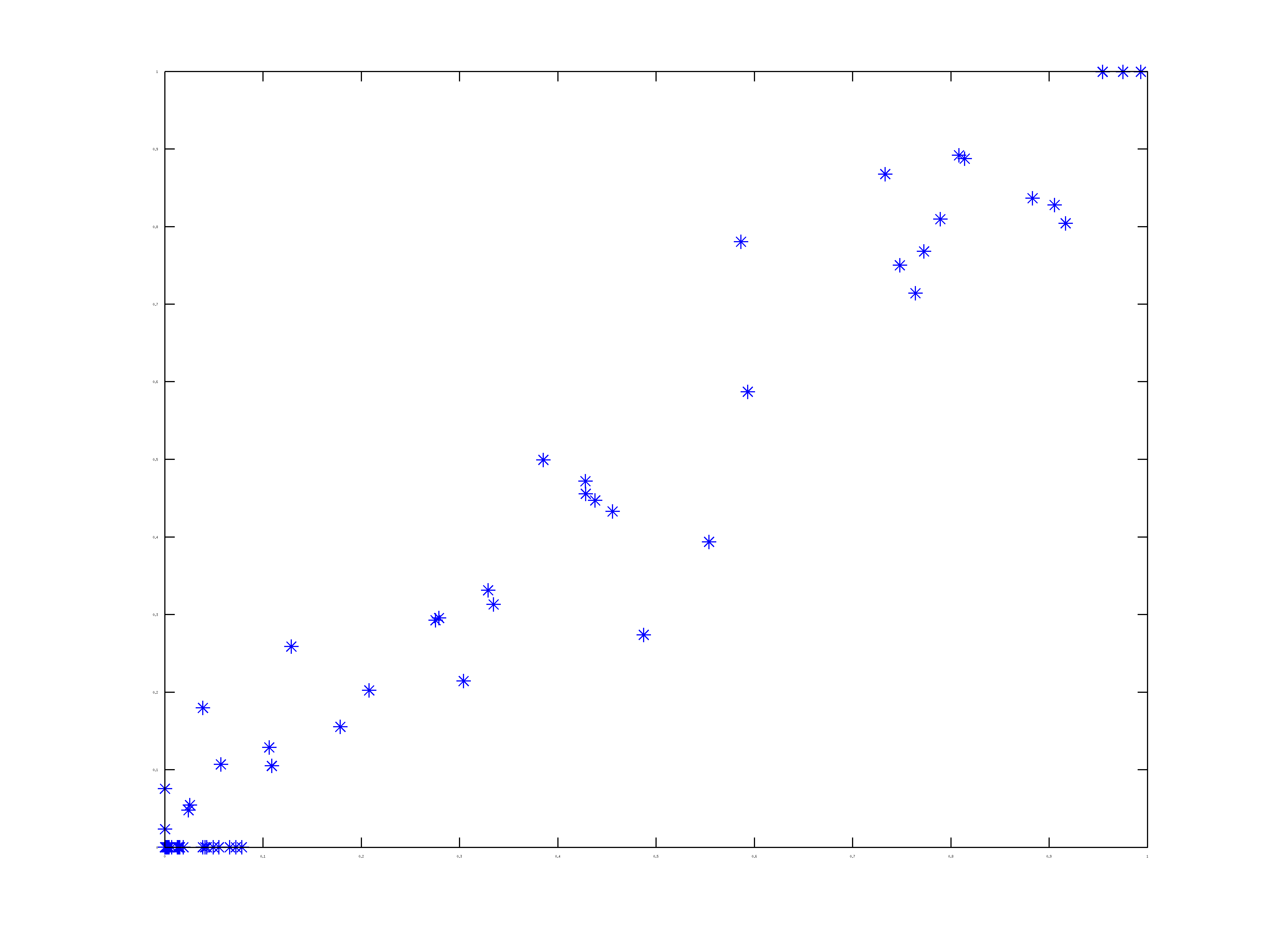}} 
    \subfigure[$\pi_4$]{\label{fig:simulation-gamma-K4-h}\includegraphics[scale=0.15]{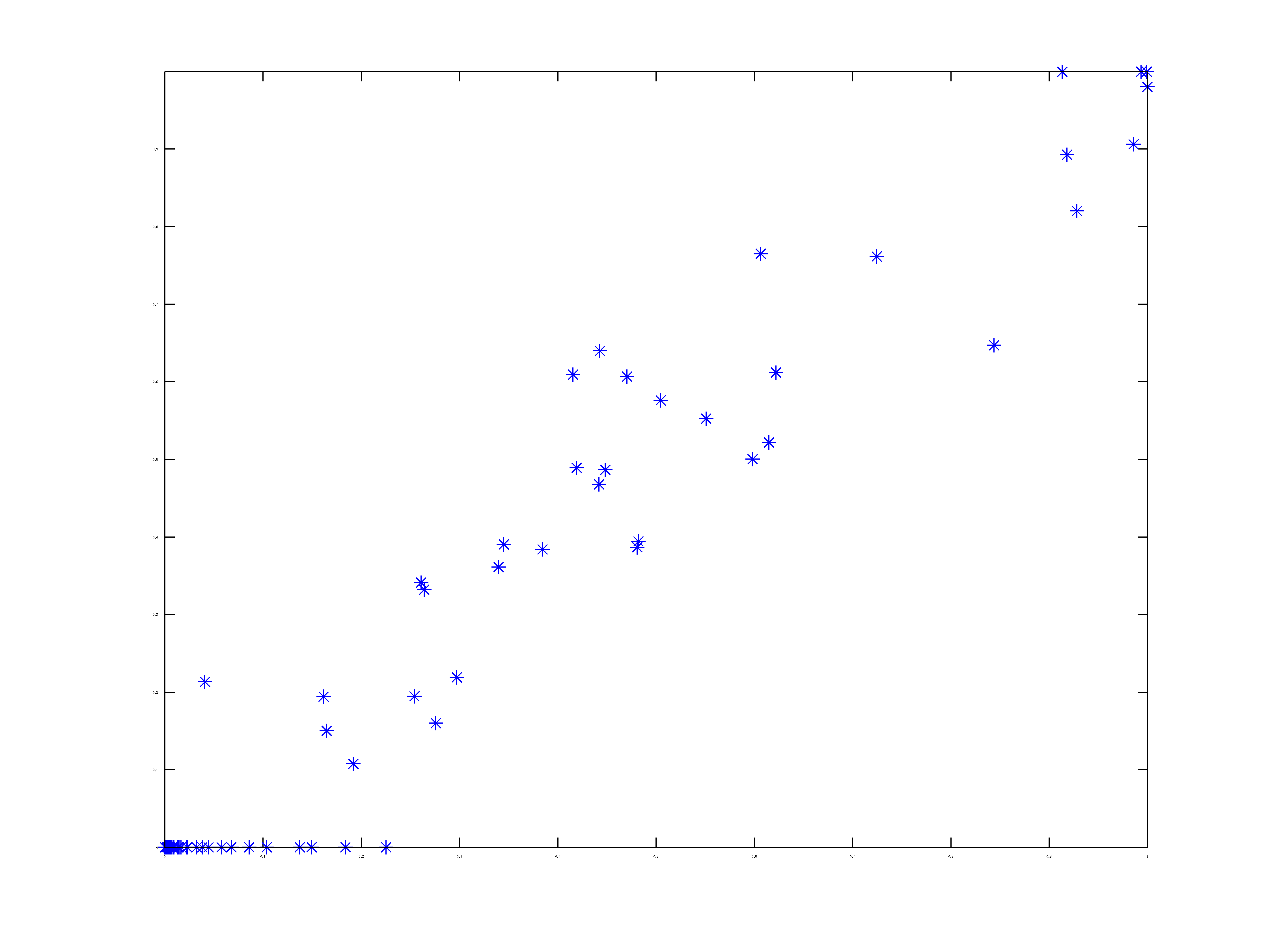}}    

  \end{center}
  \caption{True and predicted $\pi$ for two simulated examples.  Each
column corresponds to a different element of $\pi$.  True
values are on the horizontal axis. The first and the second
row show the results for $K=4, \alpha=0.05$ and $K=4,
\alpha=0.25$ respectively.
}
  \label{fig:simulation-gamma-K4}
\end{figure*}

We also verify the accuracy of predictions in $\pi$ vectors in
Figure~\ref{fig:simulation-gamma-K4}, by plotting the actual elements of
$\pi$ against the predicted values.  The majority of points lie close to a
45-degree line, indicating that the ``true'' membership probabilities are
recovered from the simulated data using the inference algorithm.
\subsection{Datasets}\label{sec:datasets}

We consider a version of the Enron email dataset provided by J. Shetty and J. Adibi~\footnote{\url{http://www.isi.edu/~adibi/Enron/Enron.htm}}.  
Their cleaned dataset consists of $252,759$ emails from $151$ Enron employees.  We further subset the data, 
focusing on all messages sent in October and November, $2001$, one of the highest-volume months.  This subset contains 
$4578$ messages between $137$ distinct employees, or an average of $16.7$
messages sent per month by each employee. 
The average message has 2.45 recipients (in any of {\tt To:, CC:, BCC:} fields). In the language of our 
paper, the sending of an email to one or more recipients is a ``transaction''.

We also present results on a transactional network extracted from \url{www.reddit.com}. 
Reddit is a social news website where users post links to content available on the Web. These postings 
can generate a series of comment chains by other users.
Reddit has topical sections called ``subreddits''. Each subreddit focuses on a topic and 
there are hundreds of them, most of which are created by users. Each post is assigned to one of the 
available subreddits by the posting user. Each post or comment is 
accompanied by other information including timestamps and voting
information. Because of the close community of users and their common
interests, this website is a great resource for research on social
networks.

We consider a transaction to be a comment or posted link and the set
of its immediate follow-up comments. The user who posted the link
or comment is the ``sender'' and all users who replied to this link
or comment are ``recipients''. The interpretation of $B$
is different from an email network.  Here,
$B_{ij}$ represents the probability that a user from group $j$
is interested in links posted by a user from group $i$.

Our dataset was derived from a crawl of the links and comments for the
top $50$ popular subreddits.  We subsetted this group, selecting the
10 most active subreddits and discarding all users with fewer than 250
submitted posts or comments. The resulting network has $248$ nodes and
$6222$ transactions.  The mean number of recipients per sender is 1.15.
We removed $500$ transactions from this data to use as a test set.  
These $500$ transactions were randomly selected from messages sent by the 10
most active nodes.

One of the features of this new dataset is the set of categories
(``subreddits''). Each new post or comment is assigned to a category.  Each user's
frequency of posting in the 10 subreddits characterizes
that user's activity.  This 10-vector can be taken
as the observed membership frequency and when normalized, as the observed
membership probability vector. This information will be
used as ``ground truth'' in section~\ref{sec:RedditClust} to calculate the
cluster performance measures developed in section~\ref{sec:Measures}.
\subsection{Exploring the Enron Dataset}\label{sec:Enron} Our
analysis of the Enron data focuses on $K=9$ groups.  Values of BIC in
Table~\ref{tb:EnronBIC} suggest this is a good group size.

Although employees have mixed membership, the membership probabilities
($\pi_i$'s) are quite focused, with half of the employees having a maximum
$\pi$ element of 0.85 or larger.  Thus much of our analysis deals with
assigning each employee to their most probable class.

\begin{figure}[htbp]
  \begin{center}
    \subfigure[Observed message frequency matrix]{\label{fig:freqmat}\includegraphics[scale=0.35]{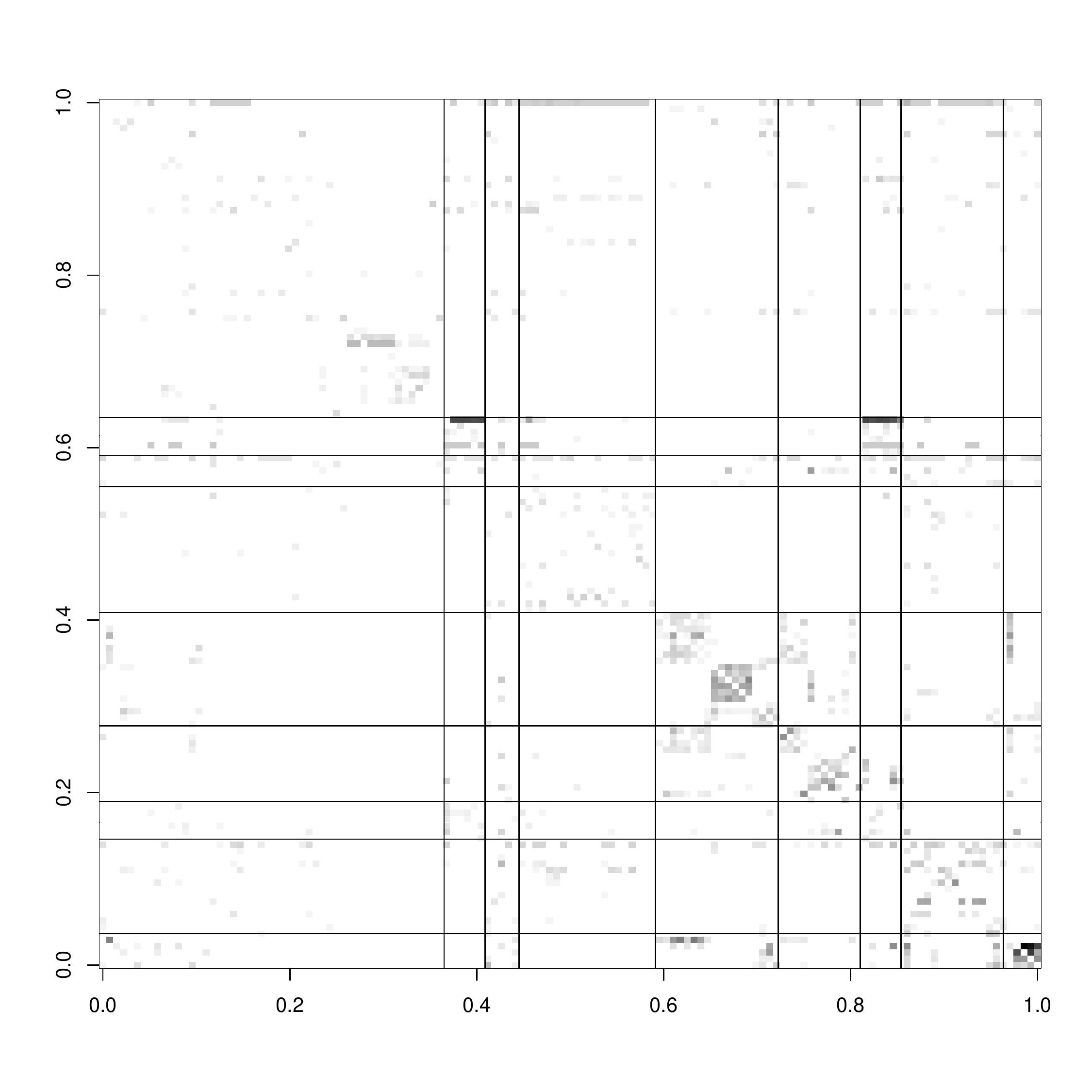}}  
    \subfigure[Predicted message frequency matrix]{\label{fig:Efreqmat}\includegraphics[scale=0.35]{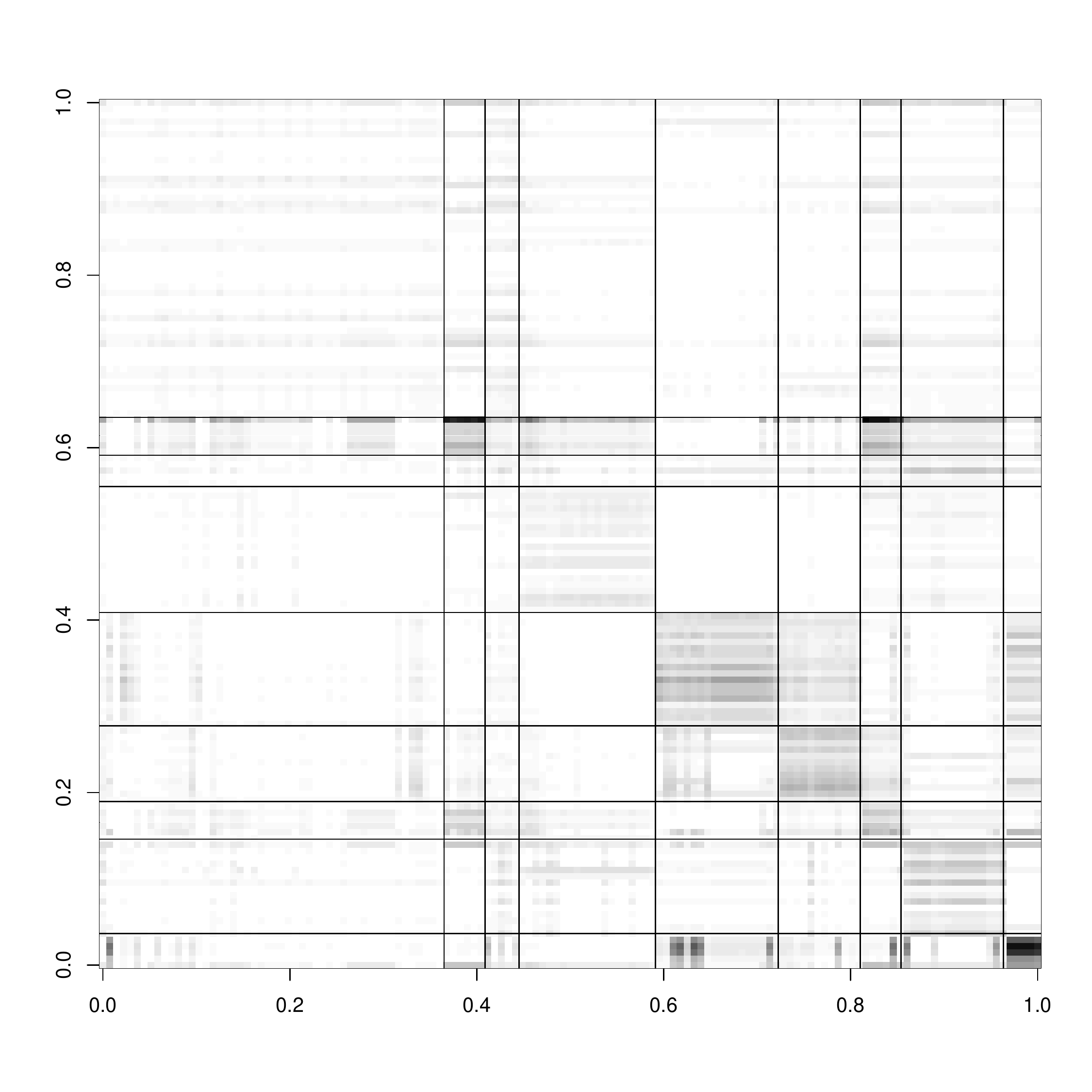}}
  \end{center}
  \caption{Enron data.  Top: Observed message frequency matrix, with rows and columns ordered by
the group ids.  Rows correspond to sender, columns to receiver.
Darker cells indicate higher frequencies.  
Lines indicate the 9 employee groups. Bottom: Predicted message frequency matrix, same layout as
Figure~\ref{fig:freqmat}. }
  \label{fig:OEfreqmat}
\end{figure}

\begin{center}
\begin{table*}[ht]
	\centering
	\begin{tabular}{r|rrrrrrrrrr}
	$K$			&	$2$	&	$3$	&	$4$	&	$5$	& 	$6$	&	$7$	&	$8$	&	$9$	&	$10$	\\ 
	\hline
	BIC ($\times 10^4$)	& $-10.1770$  	& $-8.9180$	&  $-8.8681$  	&  $-8.6310$  	&   $-8.4102$ 	&  $-8.2022$  	& $-8.1667$   	&   $-7.7530$ 	&   $-7.7735$ 	\\
	\end{tabular}
	\caption{BIC scores for different number of groups $K$ on the Enron dataset}\label{tb:EnronBIC}
\end{table*} 
\end{center}
Grouping employees by their most probable class, we present the observed
and predicted message
frequency matrices in Figure~\ref{fig:OEfreqmat}.
The groups identified by
the model appear to consists of clusters of employees who email primarily
to others in the same cluster.  The predictions in Figure~\ref{fig:Efreqmat}  
are generated by first calculating
$\Pr(j \mbox{ receives} | i \mbox{ sends}) = p_{ij} = \pi_i B \pi_j^T$
and multiplying this by the number of messages sent by employee $i$.

The predicted message frequency matrix in Figure~\ref{fig:Efreqmat}
suggests that the model is capturing some characteristics of the original
data.  The same block diagonal cells are dark (large values)
in both plots.  The same horizontal band is present within group 2 for both
observed and predicted matrices.  This band corresponds to one employee who
sent about 175 emails to the other members of group 2.  This pattern, along
with messages sent among members of group 2, appear to define this group.
Within each group, the predicted frequencies are more homogeneous
than the observed frequencies, suggesting that there remains some
subject-level behavior not represented by the model.

\begin{table*}
\begin{center}
\begin{tabular}{rrrr|rrrrrrrrr|r}
     &      &      &   &  \multicolumn{9}{|c|}{$100 \times B$} & sending  \\ 
$n_{\mbox{sent}}$ & 
$n_{\mbox{recv}}$ & 
$E(m_i)$ & 
$\hat{m}_i$ &  1 & 2 & 3 & 4 & 5 & 6 & 7 & 8 & 9 & group\\ \hline
7.9 & 21.4 & 44.1 & 50 & \fbox{1.0} & 0 & \fbox{7.7} & 0 & 0 & 0 & 1.2 & 0.6 & 0 & 1\\
28.6 & 150.8 & 9.0 & 6 & 0 & 60.2 & 2.1 & 5.4 & 0 & 0 & \fbox{77.6} & 9.9 & 0.7 & 2\\
22.5 & 43.0 & 7.8 & 5 & 0 & 0 & \fbox{0.2} & 0.2 & 1.2 & 0.6 & \fbox{4.3} &
\fbox{3.8} & 1.0 & 3\\
7.3 & 41.4 & 18.2 & 20 & 0 & 0 & 0 & 12.3 & 0 & 0 & 1.5 & 1.7 & 0.6 & 4\\
52.3 & 109.3 & 15.2 & 18 & 0 & 0 & 0.3 & 0 & 11.3 & 1.8 & 0 & 0 & 2.7 & 5\\
57.8 & 84.2 & 12.5 & 12 & 0 & 0 & 0 & 0 & 0 & 9.4 & 1.7 & 0 & 0.6 & 6\\
35 & 200.4 & 6.9 & 6 & 0 & 0.7 & 1.2 & 0 & 0 & 0.8 & 18.8 & 0.3 & 0.6 & 7\\
31.4 & 88.0 & 15.0 & 15 & 0 & 0 & 0.2 & 0 & 0.4 & 0.1 & 0.1 & 10.9 & 0 & 8\\
172.6 & 288.6 & 8.3 & 5 & 0 & 0 & 0 & 0 & 0.2 & 0 & 0.1 & 0 & 24.4 & 9\\
\end{tabular}
\caption{Enron data: summaries of the discovered groups.  The 9 
groups are ordered by row.  Boxed cells are noteworthy and discussed in the
text.
$n_{\mbox{sent}}=$ number of messages sent per employee,
$n_{\mbox{recv}}=$ number of messages received per employee,
$E(m_i)=$ predicted group size,  $\hat{m}_i=$ count of
employees whose most probable class matches this row.  Calculation of these
quantities is discussed in the text.
\label{tab:enronsummary}
}
\end{center}
\end{table*}

Table~\ref{tab:enronsummary} provides several summaries of the nine
groups.  The first three columns are calculated using probability weights
($\pi$'s) from the model.  For example, if employee $i$ sends $n_i$ messages
and has probability $\pi_{i2}$ of belonging to group 2,
then an employee in cluster 2 would send an expected
$n_{\mbox{sent}} = \sum_i\pi_{i2}n_i / \sum_i \pi_{i2}$ messages.
The activity levels vary considerably by group, as indicated by
the wide range of $n_{\mbox{sent}}$ and $n_{\mbox{recv}}$ values. The rows
of $B$ are often dominated by the diagonal element,
suggesting that most
identified groups tend to send to members of their own group.  We comment
on several interesting exceptions indicated by boxed entries 
in Table~\ref{tab:enronsummary}:
\begin{itemize}
\item Group 1 has lowest activity ($n_{\mbox{sent}}$ and $n_{\mbox{recv}}$
in row 1), and is unlikely to receive messages from anyone except another
member of group 1 (column 1 of $B$).
\item Group 1 is more likely to send to group 3 than to members of its own
group (row 1 entries).
\item Group 2 is highly likely to send to members of both group 2 and 7
(row 2 entries).
\item Group 3 has a low overall probability of sending a message, but is 
more likely to send to groups 7 and 8 than group 3 (row 3 entries).

\end{itemize}

We examine group 9 in more detail. 
Row and column 9 of $B$ 
(Table~\ref{tab:enronsummary}), indicate that group 9 
sends messages almost exclusively to other members of group 9,
and has a small but nonzero chance of 
receiving messages from most groups.  An exception is that group 9 has
negligible probability of receiving messages from groups 1 or 8.
The number of messages sent and received between possible members of 
Group 9 is displayed in Table~\ref{tab:enrongroup9}.  There are 13 nodes
that have probability of 0.25 
or more of belonging to this group ($\hat{\pi}_9$, last column of Table~\ref{tab:enrongroup9}).  A block structure is quite
evident in Table~\ref{tab:enrongroup9}:
Subgroup I 
(nodes 3, 4, 7, 9, 11, 18) and Subgroup II (nodes 17, 19, 27, 35, 57, 60, 61)
communicate primarily amongst themselves, but very little with the other subgroup.
Although it may be surprising that these two subgroups are assigned to a
single group, we note that group membership is indicated by similar {\em
sending behavior}, not necessarily the sending of messages to the same
individuals.  In this case, all nodes that have appreciable probability of
belonging to group 9 share several characteristics, namely the tendency to
send only to members of the same group, and the tendency to receive
messages from a scattering of other groups.  Inspection of $B$ suggests
that no other group has this profile.

It is also interesting to note that 3 employees in Subgroup II
(Dasovich-17, Steffes-19, and Shapiro-60), with large membership probabilities
for group 9, were all involved in ``Government relations''.

\begin{table*}
\begin{center}
\begin{tabular}{r|rrrrrr|rrrrrrr|rr}
id & 3& 4& 7&  9&11&18& 17& 19& 27&35 &57& 60& 61& oth & $\hat{\pi}_9$ \\    \hline
3 & 0& 3&14&  5& 3& 2&  0&  0&  0& 0 & 0&  0&  0& 14&0.29  \\
4 &45& 0&53& 33&43&30&  0&  0&  0& 0 & 0&  0&  0&  6&0.44  \\
7 &58&91& 0&100&48&85&  0&  0&  0& 0 & 0&  0&  0& 46&1.00  \\
9 &16& 4&42&  0& 1& 9&  0&  0&  0& 0 & 0&  0&  0& 33&0.49  \\
11&14&12&30&  5& 0& 5&  0&  0&  0& 0 & 0&  0&  0& 41&0.30  \\
18& 1& 1& 1&  0& 2& 0&  0&  0&  0& 0 & 0&  0&  0&  0&0.35  \\ \hline
17& 6& 0& 4&  0& 4& 0&  0&354& 70&41 &70&303&185& 63&1.00  \\
19& 0& 0& 0&  0& 0& 0&165&  0& 18&53 & 0&207& 37& 21&1.00  \\
27& 0& 0& 0&  0& 0& 0&  1&  0&  0& 0 &22 & 1&  4&258&0.30  \\
35& 0& 0& 0&  0& 0& 0&  4&  1&  4& 0 & 0 & 4&  3& 42&0.26  \\
57& 0& 0& 0&  1& 0& 0& 27&  0& 17& 0 & 0 & 0&  0& 87&0.35  \\
60& 0& 0& 0&  0& 0& 0& 57& 58&  9& 2 & 0 & 0& 68& 14&1.00  \\
61& 0& 0& 0&  0& 0& 0& 12&  9&  8& 0 & 0 &27&  0&103&0.88  \\ \hline
oth  &76&24&96& 10&22& 5&  1& 16&107&41&305 &11& 48&   &      
\end{tabular}
\end{center}
\caption{Enron data:
Observed message frequencies between the 13 employees with probability
$\geq 0.25$
of belonging to group 9.  Rows are senders, columns are 
recipients.  Lines between nodes 18 and 17 indicate two subgroups
displaying block structure. 
Rows and columns marked ``oth'' are an aggregation of all other
nodes.  For example, node 4 sends 45 messages to
node 3, and 6 messages to ``other'' nodes.
$\hat{\pi}_9$ is the estimated probability of belonging to group 9.
}
\label{tab:enrongroup9}
\end{table*}
\subsection{Exploring the Reddit Dataset}\label{sec:Reddit}
\begin{table*}
\begin{center}
\begin{tabular}{rrrr|rrrrrr|r}
     &      &      &   &  \multicolumn{6}{|c|}{$100 \times B$} & sending  \\ 
$n_{\mbox{sent}}$ & 
$n_{\mbox{recv}}$ & 
$E(m_i)$ & 
$\hat{m}_i$ &  1 & 2 & 3 & 4 & 5 & 6 &  group\\ \hline
10.5 & 10.8 & 80.1 & 96 & 0 & 0.9 & 1.1 & 0.1 & 0 & 1.5 & 1\\
17.7 & 19.3 & 51.7 & 43 & 0 & 0.6 & 3.4 & 0.1 & 0.1 & 1.2 & 2\\
89.5 & 119.9 & 7.9 & 5 & 0 & 0.5 & 2.5 & 0.2 & 1.7 & 0.2 & 3\\
22.5 & 23.9 & 37.4 & 39 & 0 & 0.3 & 6.1 & 0.2 & 1.4 & 0.1 & 4\\
42.3 & 48.3 & 28.5 & 29 & 0 & 0 & 5.0 & 0.1 & 2.6 & 0 & 5\\
28.5 & 35.3 & 42.3 & 36 & 0 & 0.5 & 0.8 & 0 & 0 & 1.9 & 6\\
\end{tabular}
\caption{Reddit data: Summaries of the discovered groups.  See
caption for Table~\ref{tab:enronsummary} for explanation of entries.
\label{tab:redditsummary}
}
\end{center}
\end{table*}
As with the Enron data, we present results of an analysis with a specific
$K$.  Following this analysis, we also examine predictive performance using
the clustering and link prediction metrics developed in
Section~\ref{sec:Measures}.

Table~\ref{tab:redditsummary} shows the $B$ matrix and other summaries for
a model with $K=6$ clusters.  Summaries are the same as for the Enron data in
Table~\ref{tab:enronsummary}.
Entries of $B$ are considerably smaller than in
the Enron case.  This is due to the larger number of nodes, the smaller
number of recipients per message (1.15 compared to 2.45) and the more
``open'' nature of discussion forums compared to email communication.
The large non-diagonal entries of $B$
suggest that groups are defined in terms of {\em between} group
communications as much as {\em within}.  Group 1 ``receives'' no messages,
which implies that a member of this group will make posts to a subreddit
(i.e., ``send'' a message),
but will not follow up on another post.  Group 4 has low ``receive'' traffic as
well.  Group 3 is interesting in that it constitutes just a few members (5
nodes have this as their most probable class) who
have very high volume of posts.
\begin{table}
\begin{center}
\begin{tabular}{r|rrrrrr}
sending & \multicolumn{6}{c}{receiving group}\\
group & 1 & 2 & 3 & 4 & 5 & 6 \\ \hline
1 & 0 & 0.463 & 0.088 & 0.028 & 0.013 & 0.620\\
2 & 0 & 0.293 & 0.267 & 0.050 & 0.018 & 0.523\\
3 & 0 & 0.261 & 0.195 & 0.088 & 0.478 & 0.105\\
4 & 0 & 0.135 & 0.482 & 0.090 & 0.412 & 0.056\\
5 & 0 & 0.009 & 0.393 & 0.030 & 0.752 & 0.010\\
6 & 0 & 0.233 & 0.061 & 0.017 & 0.007 & 0.800\\
\end{tabular}
\end{center}
\caption{Reddit data: Group-size-weighted $B$.
Expected number of recipients in each group, for a message sent by
a single member of ``sending group''.\label{tab:redditErecip}}
\end{table}

One problem with the $B$ matrix is that it does not convey information on
the expected number of recipients.  For example, there is a
relatively large probability (0.061) that a message sent from a member of
group 4 will be received by a member of group 3.  However, there are only
an expected 7.9 nodes belonging to group 3, implying that the expected
number of recipients would be $0.061 \times 7.9 = 0.48$, or about half a
node.  Such a calculation can be carried out for the entire $B$ matrix by
multiplying column $j$ by the
expected number of members in group $j$, generating a group-size-weighted $B$.
The results are displayed in
Table~\ref{tab:redditErecip}.  We see that group 3 now appears less
``active'' since the expected number of recipients is smaller.  Groups 5
and 6 have large entries on the diagonal of the group-size-weighted $B$,
suggesting they are most likely to generate posts that are responded to by
members of their own group.  In contrast, posts from group 4 are much more
likely to generate responses from members of other groups, such as 3 and 5.
\subsection{Link Prediction Results for Reddit}\label{sec:RedditLinkPrediction}
We compare our method with the MMSB model and a hierarchical clustering method applied to the symmetrized version of the transaction counts, e.g. Figure~\ref{fig:ExampleData}(b).

For the problem of link prediction, we use the performance measure
developed in~\cite{NallapatiCohenAhmedXing2008}. It focuses on how well a
method ranks the true recipients. It uses the value of the rank at $100\%$
recall.  A small rank indicates that the model identifies all true
recipients before many non-recipients are identified.
For our model ranks are generated using $\Pr(j \mbox{ receives} |
i \mbox{ sends}) = p_{ij} = \pi_i B \pi_j^T$.
For each message, we
rank the nodes based on their predicted probability of being recipients
of the message. We pick the rank of the last predicted recipient as
performance measure for the message. The overall performance will be
the average of individual performances for all messages.

Direct comparisons with link prediction are immediately possible with our
model and the MMSB model, since both predict the probability of a link or
transaction between two nodes.  For hierarchical clustering, we must
develop a similar prediction.  A crude version of the $B$ matrix can be
constructed using cluster labels from hierarchical clustering, and counting
the number of messages sent and received by nodes with each label
combination.  

\begin{figure}[htp]
  \begin{center}
    \includegraphics[scale=0.35]{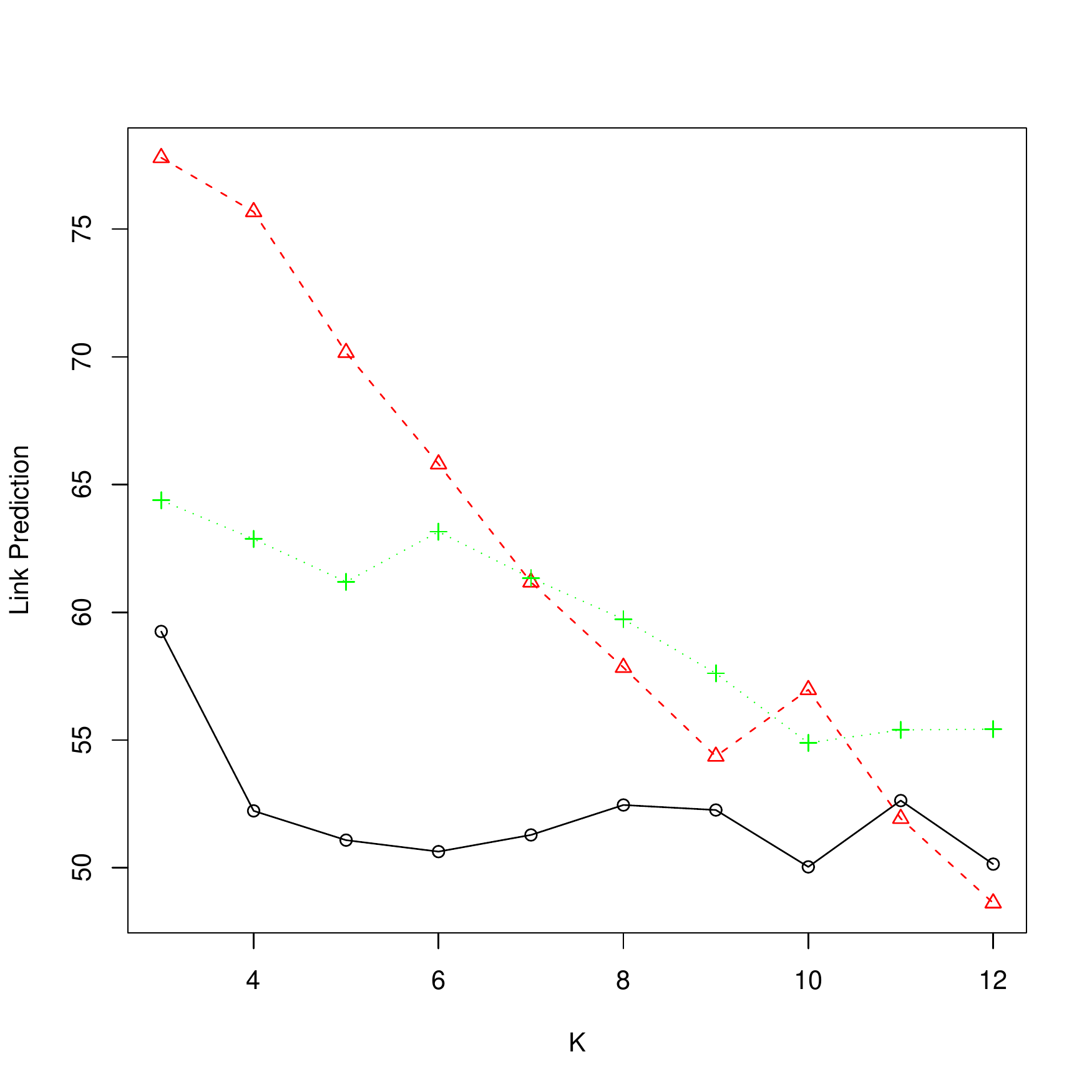}
  \end{center}
  \caption{Reddit data: Link prediction results. \textbf{black} is our model, \textbf{\textcolor{red}{red}} 
	    is the MMSB model and \textbf{\textcolor{green}{green}} is the hierarchical clustering algorithm.}
  \label{fig:LinkPredictionResults}
\end{figure}

Fig.~\ref{fig:LinkPredictionResults} shows the results for the three
methods on the Reddit test dataset.  Our method produces significantly
better (i.e. lower) scores with as few as 4 groups.
\subsection{Clustering Results for Reddit\label{sec:RedditClust}}
\begin{figure*}[htp]
  \begin{center}
    \subfigure[Precision]{\label{fig:redditPrecision}\includegraphics[scale=0.3]{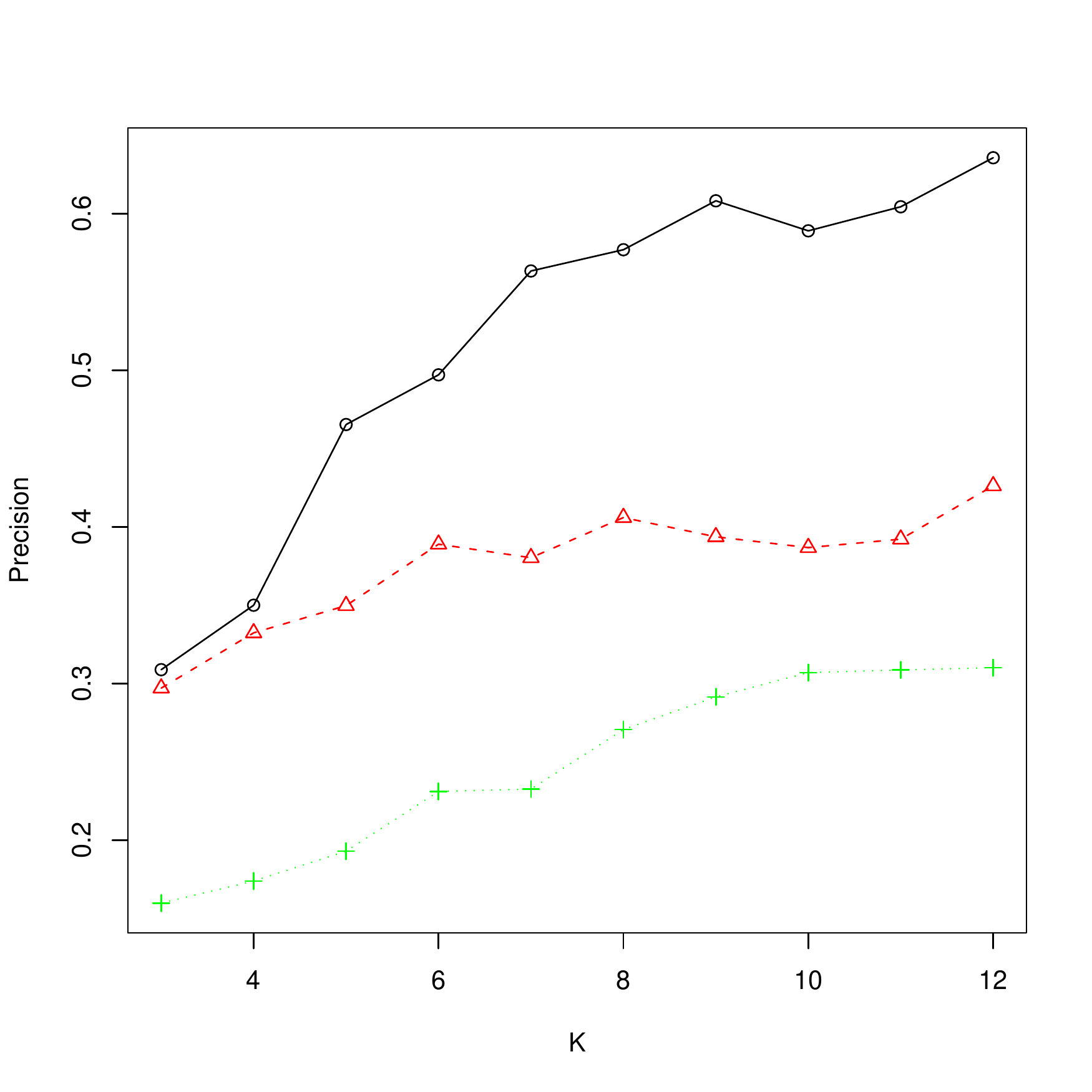}}  
    \subfigure[Recall]{\label{fig:redditRecall}\includegraphics[scale=0.3]{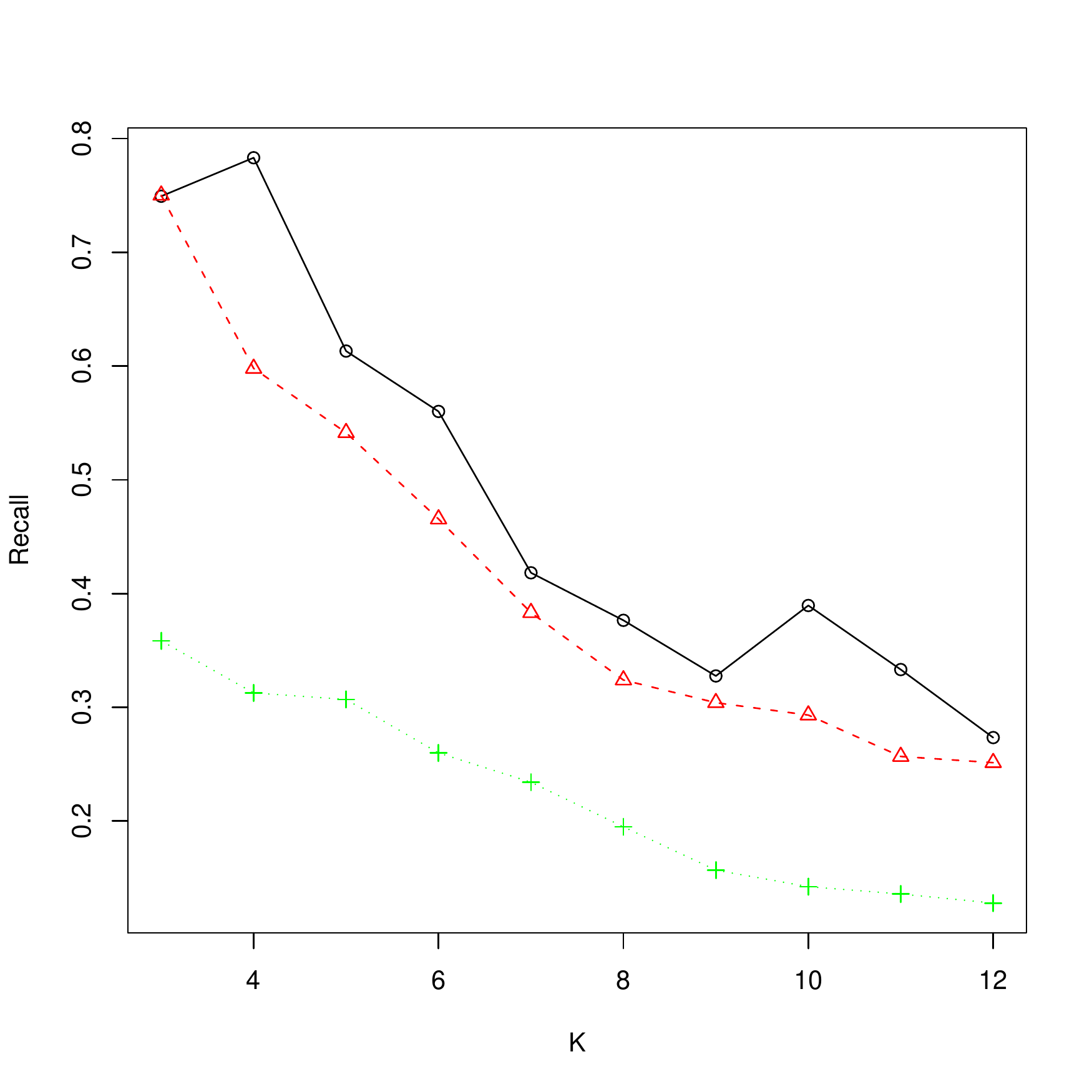}}
    \subfigure[F-Measure]{\label{fig:redditFMeasure}\includegraphics[scale=0.3]{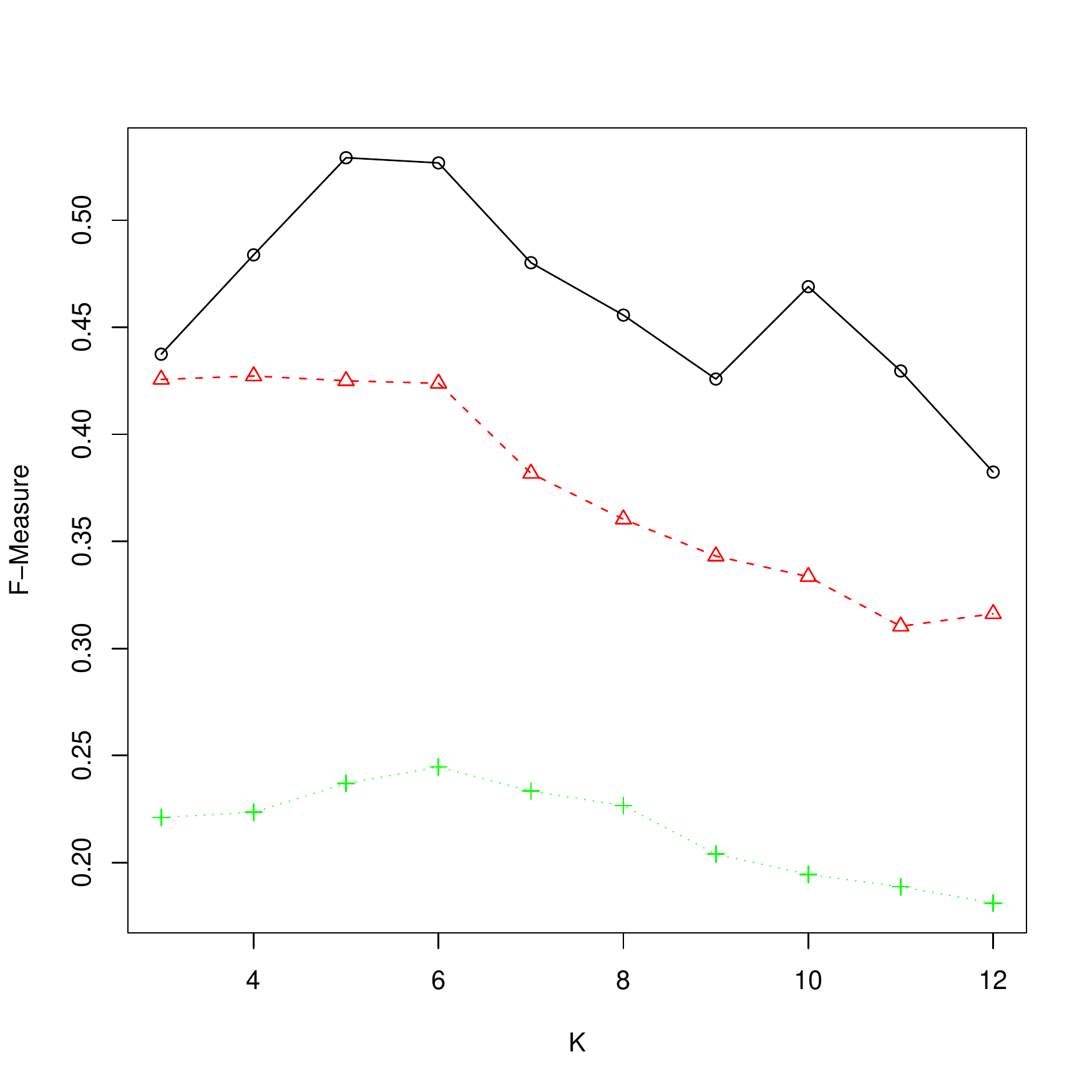}}
  \end{center}
  \caption{Reddit data: Clustering results. \textbf{black} is our model, \textbf{\textcolor{red}{red}} 
	    is the MMSB model and \textbf{\textcolor{green}{green}} is the hierarchical clustering algorithm.}
  \label{fig:RedditClusteringResults}
\end{figure*}
As in the previous section, we compare our method with the MMSB model
and a hierarchical clustering method mentioned in Section~\ref{sec:RedditLinkPrediction}

The availability of an observed mixed membership vector (based on subreddit
posting frequencies) enables us to quantify clustering accuracy in the
reddit dataset.  During training, the observed mixed membership vector is
ignored.  Using the test set described in section~\ref{sec:datasets} and
the performance measures in section~\ref{sec:Measures}, we can measure
clustering accuracy.

In Figure~\ref{fig:RedditClusteringResults}, we compare precision, recall
and F-measure for our method with values obtained for the MMSB model and
a simple hierarchical clustering model.  In the MMSB model, a threshold
of 1 transaction was used to convert transaction counts to binary data.
A hierarchical clustering method was applied to a matrix of send/receive
frequencies to generate class labels for each node.  The MMSB model and
our model both produce mixed memberships.  The hierarchical clustering
method produces hard classifications.  It appears that our method's
superior performance may be due to the mixed memberships and the ability to
utilize co-recipient and message frequency information.  Hierarchical
clustering does not produce mixed membership, and the MMSB model cannot use
co-recipients or message frequencies.

\section{Conclusions }
The key innovations of our model are the ability to probabilistically model
transactional data with multiple recipients, the generalization of
criteria for group membership to include communication patterns with other
groups, and the development of a model in which individuals can belong to
multiple groups.  The variational inference
algorithm is efficient for large networks, and can accurately recover
network structure.  The real data examples indicate that the model can
extract interesting information from network data, and that it is
competitive in discovering mixed memberships and in predicting
transactions. We proposed a novel performance measure for comparing soft clustering results.

One issue not yet discussed is scalability of the
algorithm. We studied how the TMMSB model scales with respect to the
number of nodes ($M$), 
transactions ($N$) and cluster size ($K$). We fit the model to a set of simulated networks
with $50 \leq M \leq 400$ nodes, $500 \leq M \leq 6000$ transactions and $3 \leq K \leq 10$ clusters.
Sufficient combinations were explored to enable estimation of a model relating time to these parameters: 
$$\mbox{\tt time} \propto M^{1.36}N^{1.36}K^{2.71},$$
This multiplicative model indicates that complexity grows much more rapidly with
respect to $K$ than with respect to $N$ or $M$. Computation times varied from 3 minutes with $(M,N,K)=(50,500,3)$ to 3 days with $(M,N,K) = (400, 6000, 10)$ or $(300,4000,7)$.  All computations were carried out on a Linux-based cluster with AMD Opteron
CPUs (clock speeds varying from 2.6 - 3.0GHz). The fitted model, a linear
regression of log time on logged $M,N,K$, had a high $R^2$, about 98.4\% of
variation in log time being described by a main effects model.

Our model could be extended in several ways.  One shortcoming of 
the Bernoulli model for recipients is that it permits transactions
with no recipients, an impossible outcome in email transactions.
Extensions that exclude such null transactions might capture additional
structure.  Other transaction information such as
timestamps, headers and content could be incorporated as covariates.
Time-varying versions of this model could be used to discover changes in
group membership and activity. This could include either varying
memberships ($\pi$'s) or changing numbers of groups (varying $K$). The
simplest way to fit such a model would be to partition the time axis into
intervals, and fit a separate model in each interval.  More complex models
could be considered.  For example the MMSB model was extended to time
intervals by \cite{Huh.Fienberg.2008}, and dependence between estimated parameters was introduced
via a Markov assumption, in which the parameter values were dependent between
one time interval and the next.

Another extension would be to associate different group
structure with the sending and receiving of messages.  In the current
model, the sending and receiving behavior is governed by the same
groups.  The additional structure would allow separation of a distribution
of ``message topics'' from the distribution of group memberships of nodes.

An earlier version considered had additional structure, in which each
transaction had a group associated with it, as well as the $M$ nodes.   In
this model, a group label was assigned to the transaction, and each node
drew its group label.  The sending node was then selected from those nodes
whose current group matched the group label of the transaction.  The TMMSB
model described here dispenses with the additional step, simply choosing
the sender from all possible nodes, and then having the message label
corresponding to the group label of the sender.  The additional structure
would allow separation of a distribution of ``message topics'' from the
distribution of group memberships of nodes.  It does however assume that
message groups and node groups have a 1:1 correspondence (same number and
interpretation of categories), which might not be realistic.
It is unclear whether
there is sufficient information in the data to allow estimation of this
additional structure, or even whether that structure serves much practical
purpose.

\section*{Acknowledgements}
This research was supported by the Natural Sciences and Engineering Research Council of Canada (NSERC), the Mathematics of 
Information Technology and Complex Systems (MITACS) and 
the Atlantic Computational Excellence network (ACEnet). The authors would like to thank Edo Airoldi for helpful discussions.

\bibliographystyle{abbrv}

\bibliography{TMMSB}

\end{document}